\PassOptionsToPackage{table,xcdraw}{xcolor}
\documentclass{article} % For LaTeX2e
\usepackage{iclr2027_conference,times}

\renewcommand{\iclrruler}[1]{}

\usepackage{amsmath,amsfonts,bm}

\def\eqref#1{equation~\ref{#1}}
\def\1{\bm{1}}

\DeclareMathAlphabet{\mathsfit}{\encodingdefault}{\sfdefault}{m}{sl}
\SetMathAlphabet{\mathsfit}{bold}{\encodingdefault}{\sfdefault}{bx}{n}

\usepackage{url}
\usepackage{graphicx}
\usepackage{wrapfig} % Half-width motivation figure with body text alongside.
\usepackage{caption}
\usepackage{booktabs}
\usepackage{amsmath}
\usepackage{amssymb}
\usepackage{amsfonts}
\usepackage{array}
\usepackage{multirow}
\usepackage{xcolor}
\usepackage{xspace}
\usepackage{adjustbox}
\usepackage{microtype}
\usepackage[colorlinks=true,linkcolor={black!62!blue},citecolor={black!62!blue},urlcolor={black!62!blue}]{hyperref}

\definecolor{gaincolor}{HTML}{3F9DA2}
\definecolor{placeholdergray}{gray}{0.6}
\definecolor{oursgray}{gray}{0.90}
\definecolor{oursbg}{gray}{0.92}
\definecolor{gainbg}{gray}{0.96}
\definecolor{oursrow}{RGB}{235,243,252}  % light-blue background marking +Ours cells
\newcommand{\gainpos}[1]{\cellcolor{gainbg}\textcolor{gaincolor}{#1}}
\newcommand{\ourscell}[1]{\cellcolor{oursbg}{#1}}
\newcommand{\Up}{\ensuremath{\uparrow}}
\newcommand{\Down}{\ensuremath{\downarrow}}

\newcommand{\up}{\ensuremath{\uparrow}}
\newcommand{\down}{\ensuremath{\downarrow}}

\newcolumntype{O}{>{\columncolor{oursrow}\bfseries}c}
\newcolumntype{M}[1]{>{\centering\arraybackslash}m{#1}}
\def\name{\emph{\mbox{SR$^{2}$-Net}}\xspace} % kept on one line

\title{Correcting Spectra Outside the Backbone:\\ A Model-Agnostic Rectifier for Hyperspectral\\ Image Super-Resolution}

\author{Ji-Xuan He\textsuperscript{1,*}\quad Guohang Zhuang\textsuperscript{2,*}\quad Bo Junge\textsuperscript{1}\quad Lingchen\textsuperscript{1}\\
\bfseries Tingyi Li\textsuperscript{2}\quad Yanan Qiao\textsuperscript{1,\textdagger}\quad Miaomiao Cai\textsuperscript{3}\quad Junfeng Fang\textsuperscript{3}\\
\normalfont\footnotesize\textsuperscript{1}Xi'an Jiaotong University\quad
\textsuperscript{2}Hefei University of Technology\quad
\textsuperscript{3}National University of Singapore\\
\normalfont\footnotesize\textsuperscript{*}Equal contribution.\quad
\textsuperscript{\textdagger}Corresponding author.}

\iclrfinalcopy
\hypersetup{
  pdftitle={Correcting Spectra Outside the Backbone: A Model-Agnostic Rectifier for Hyperspectral Image Super-Resolution},
  pdfauthor={Ji-Xuan He, Guohang Zhuang, Bo Junge, Lingchen, Tingyi Li, Yanan Qiao, Miaomiao Cai, Junfeng Fang}
}
\begin{document}

\maketitle
\lhead{Preprint}

\begin{abstract}
\looseness=-1 Hyperspectral image super-resolution (HSI-SR) aims to recover spatial detail while preserving the spectral shape on which quantitative analysis relies. Recent HSI-SR methods, from repurposed RGB super-resolution backbones to dedicated spectral-spatial architectures, have greatly improved spatial reconstruction. However, overlooking the compact spectral structure of hyperspectral data leaves residual spectral errors, while binding the spectral treatment to each architecture forces it to be rebuilt for every new backbone. Yet the low-dimensional structure of spectra belongs to the data, not to any backbone. Backbones differ in the errors they leave, but not in the structure of the true spectra. One rectifier design can therefore serve any backbone. Building on this insight, we propose the \textbf{S}pectral \textbf{R}ectification \textbf{S}uper-\textbf{R}esolution Network (\textbf{\name}), a model-agnostic rectifier that needs nothing from the backbone but its output, leaves its internal architecture untouched, and is trained per backbone. \name follows an \emph{enhance-then-rectify} pipeline in which Hierarchical Spectral-Spatial Synergy Attention (\textbf{H-S$^{3}$A}) reinforces cross-band interactions, while Mode-Constrained Rectification (\textbf{MCR}) confines the correction to a learned compact spectral subspace. A degradation-consistency constraint further ties the output to the observed low-resolution input. Experiments with five backbones spanning CNN, Transformer, and diffusion families show that one fixed configuration improves spectral fidelity in every reported setting while preserving or improving spatial quality. Averaged over thirty in-domain settings,~\name removes 22.6\% of the residual spectral error at a backbone-independent cost of 0.048M parameters.
\end{abstract}

\section{Introduction}
A hyperspectral image~(HSI) records tens to hundreds of contiguous spectral bands at each pixel, and this dense spectrum is what makes it useful from remote sensing to biomedical imaging~\citep{aburaed2023review,gao2025msfmamba,wang2021review,yang2025hyperspectral}. Sensors trade spatial resolution for spectral resolution, so a hyperspectral cube is usually spectrally rich but spatially coarse, and Hyperspectral Image Super-resolution~(HSI-SR) fills in the missing spatial detail. Unlike in three-channel natural-image SR~\citep{liang2021swinir,chen2023activating}, the recovered bands must stay radiometrically faithful, since once the spectral shape or its absorption features drift, the quantitative analysis that motivates hyperspectral imaging no longer holds~\citep{kotwal2025hyperspectral}.

\begin{wrapfigure}{r}{0.50\textwidth}
    \centering
    \includegraphics[width=\linewidth]{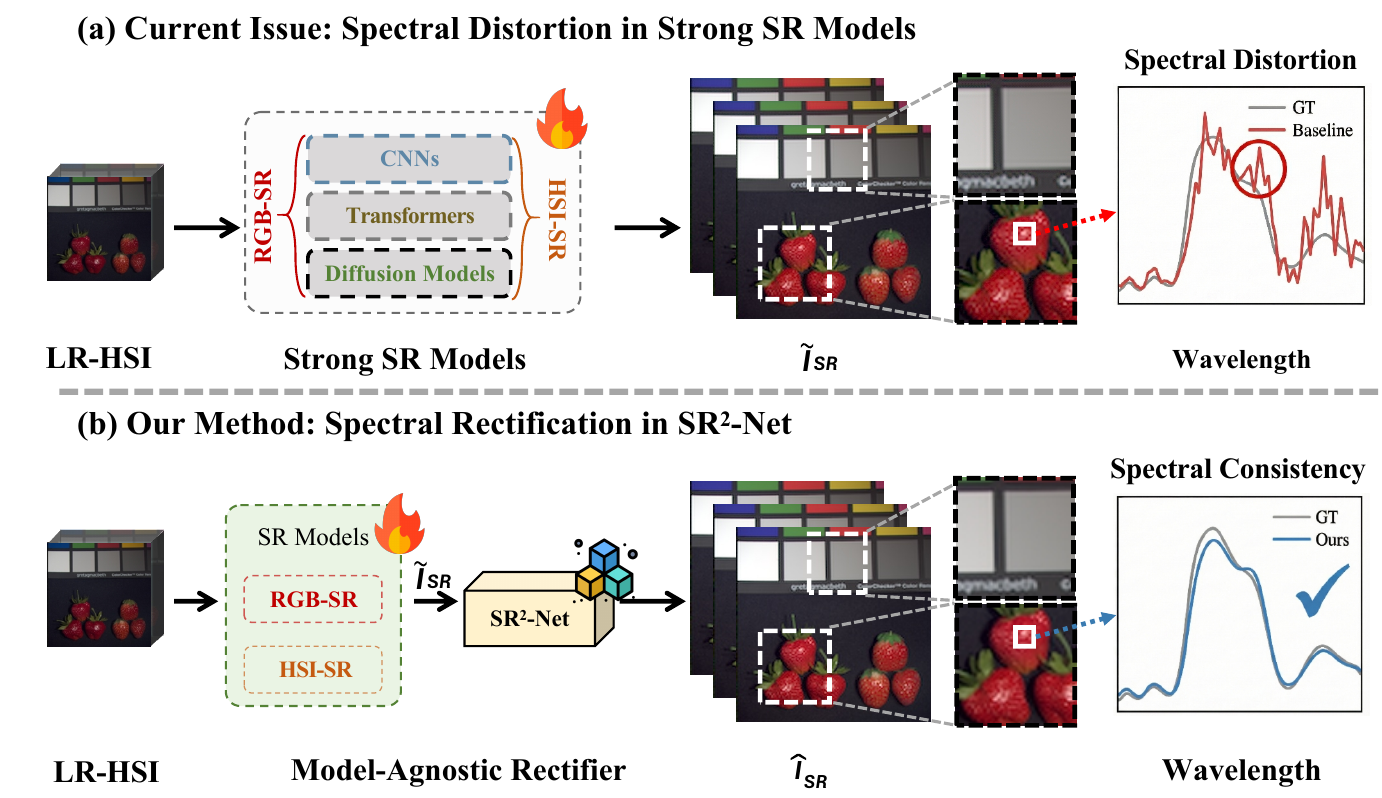}
    \caption{Motivation. (a) Strong SR models, general-purpose and HSI-specific alike, introduce spectral distortion during reconstruction; (b) the model-agnostic~\name rectifier restores \mbox{spectral consistency.}}
    \label{fig1}
\end{wrapfigure}

\looseness=-1 Current HSI-SR methods follow two directions. One repurposes mature RGB-SR backbones such as RCAN~\citep{zhang2018image} and SwinIR~\citep{liang2021swinir}, which recover spatial texture well but treat the bands as ordinary channels. The other builds HSI-specific architectures such as SSPSR~\citep{jiang2020learning} and ESSAformer~\citep{zhang2023essaformer} with dedicated spectral-spatial interaction mechanisms~\citep{wu2024unsupervised,zhang2025cassnet}. Although both recover spatial detail well, the spectra they reconstruct still carry residual error, drifting from the ground-truth shape even where the images in Figure~\ref{fig1} look spatially plausible~\citep{he2024connecting}. The root lies in a premise the two directions share, namely that spectral fidelity is a capability every architecture has to build in for itself. That premise leads to two challenges.
% Keep the motivation figure at page 2 top, with text wrapping on the left.
\newpage
\noindent
\textbf{\textit{1) Overlooking the compact spectral structure of hyperspectral data leaves the reconstructed spectral shape unconstrained.}} Neighboring bands sample a continuous reflectance response, so a scene's spectra span only a few principal directions. Few methods turn this regularity into a constraint. RGB-derived backbones are tuned for spatial texture and impose nothing on the multi-band axis, whereas most HSI-specific models, despite their dedicated interaction mechanisms, still learn cross-band relations implicitly, so nothing in the reconstruction states what a valid spectrum looks like. The spectral axis is then corrected only as a by-product of minimizing a pixel-wise loss, and band-to-band error survives.
\textbf{\textit{2) Binding the spectral treatment to a specific architecture forces it to be rebuilt for every new backbone.}} Whatever spectral treatment an architecture does have, including the explicit low-dimensional priors of EigenSR~\citep{su2025eigensr} and abundance-domain reconstruction~\citep{hu2024exploring}, is embedded in its own layers or solving domain, so the mechanism built for one cannot be reused by the next. Adopting a stronger spatial backbone means solving spectral fidelity again from scratch, and the effort never accumulates across the field.

Both challenges follow from the same premise, and that premise need not hold. The low-dimensional structure of spectra~\citep{DBLP:journals/tgrs/ChenZHZJH24} belongs to the data, not to any backbone. Backbones differ in the spectral errors they leave behind, but not in the structure of the true spectra. The missing constraint therefore belongs in a module of its own, stated explicitly rather than learned implicitly and placed outside rather than rebuilt inside each backbone, so that one rectifier \mbox{design serves any backbone.}

\looseness=-1 In this paper, we propose the \textbf{S}pectral \textbf{R}ectification \textbf{S}uper-\textbf{R}esolution Network (\name), a model-agnostic rectifier that needs nothing from the backbone but its output: it leaves the internal architecture untouched and is trained per backbone, even on a frozen one. The second challenge is answered by construction, since the same fixed design and configuration attach to CNN, Transformer, and diffusion backbones alike with no redesign. The first challenge is answered by a two-stage \emph{enhance-then-rectify} strategy, whose stages supply the two sides of the missing constraint. Hierarchical Spectral-Spatial Synergy Attention (H-S$^{3}$A) first makes cross-band interaction explicit, strengthening the dependencies that the backbone learns only weakly. Mode-Constrained Rectification (MCR) then confines the correction to a few learned spectral modes, so that it reshapes a spectrum along those directions rather than band by band. A degradation-consistency constraint keeps that freedom anchored, requiring the reconstructed high-resolution (HR) image, once degraded, to reproduce the low-resolution (LR) input. Our main contributions are as follows:
\begin{itemize}
    \item We trace the residual spectral error of HSI-SR to the premise that spectral fidelity must be built into each architecture, and recast spectral correction as a backbone-independent problem grounded in the low-dimensional spectral structure of the data.
    \item We realize this as~\name, an \emph{enhance-then-rectify} rectifier in which H-S$^{3}$A makes cross-band interaction explicit, a deliberately simple MCR bottleneck confines the correction to a learned compact spectral subspace, and an auxiliary degradation-consistency constraint keeps the output faithful to the LR measurement, all without modifying the \mbox{backbone's internal architecture.}
    \item We validate~\name on five CNN, Transformer, and diffusion backbones, where one fixed configuration improves spectral fidelity in every reported setting and removes 22.6\% of the in-domain residual spectral error on average for 0.048M extra parameters. With the backbone frozen, it still recovers 76\% of the jointly trained gain, and controlled comparisons show that the gain is not a capacity effect, since widening the MCR bottleneck to the band count gives part of it back.
\end{itemize}

\section{Related Work}

\subsection{Image and Hyperspectral Super-Resolution}
Deep SR architectures have steadily improved spatial detail recovery, and strong backbones are now available across families, including CNN-based RCAN~\citep{zhang2018image}, Transformer-based SwinIR~\citep{liang2021swinir} and HAT~\citep{chen2023activating}, the state-space MambaIR~\citep{guo2024mambair}, and diffusion-based ResShift~\citep{yue2023resshift} and DiffBIR~\citep{lin2023diffbir}. Hyperspectral image super-resolution~(HSI-SR) inherits these designs but must also preserve band-to-band continuity, a requirement it shares with recovering a full spectral cube from a single RGB image~\citep{mei2022learning,wang2024hyperspectral,wang2024whanet}. A first group treats spectral bands as channels and trains general-purpose backbones on hyperspectral cubes, inheriting strong spatial reconstruction but constraining the spectral axis only implicitly. A second, larger group designs spectral-spatial coupled architectures, exemplified by SSPSR~\citep{jiang2020learning} and ESSAformer~\citep{zhang2023essaformer} and by remote sensing work that jointly reconstructs spatial and spectral content~\citep{DBLP:journals/tgrs/ZhangHWL21,DBLP:journals/tgrs/LiG22,DBLP:journals/tgrs/ZhuDDZ22,DBLP:journals/tgrs/ChenZZ23a,DBLP:journals/tgrs/ZhangQWTHHL25,dong2023model}, while self-supervised and unsupervised schemes~\citep{DBLP:journals/tgrs/HongYLMYC23,wu2024unsupervised} and diffusion priors~\citep{miao2023dds2m,pang2024hir} reduce reliance on paired data. Across these directions the spectral treatment remains architecture-specific, and residual spectral errors persist.

\subsection{Spectral Fidelity and Model-Agnostic Refinement}
\looseness=-1 A complementary direction targets spectral fidelity directly rather than through architecture alone. One group enforces spectral-consistency or degradation-consistency objectives so that reconstructions stay compatible with the observed LR input~\citep{wang2022recurrent,dong2023model}. A second group makes that structure explicit inside the reconstruction, solving in the abundance domain~\citep{hu2024exploring}, bridging a pre-trained RGB backbone through eigenimages~\citep{su2025eigensr}, or learning a low-rank prior~\citep{wang2023hyperspectral}, each tied to one architecture. Classical post-hoc corrections such as iterative back-projection are hand-crafted, and a learned refiner such as the two-stage DualSR~\citep{li2022dualsr} sharpens spectra with a spectral-angle-constrained back-projection co-designed with its own coarse network. Model-agnostic spectral correction has also been studied for RGB-to-spectral reconstruction, such as the analytic projection of \citet{lin2020physically}, which needs no training but does no spatial reasoning. In contrast to these co-designed or fixed corrections,~\name is one learned rectifier design that attaches unchanged to CNN-, Transformer-, and diffusion-based backbones for spatial HSI-SR.

\section{Method}
\subsection{Overview} % Overview

\begin{figure}[!t]
  \centerline{\includegraphics[width=1.0\linewidth]{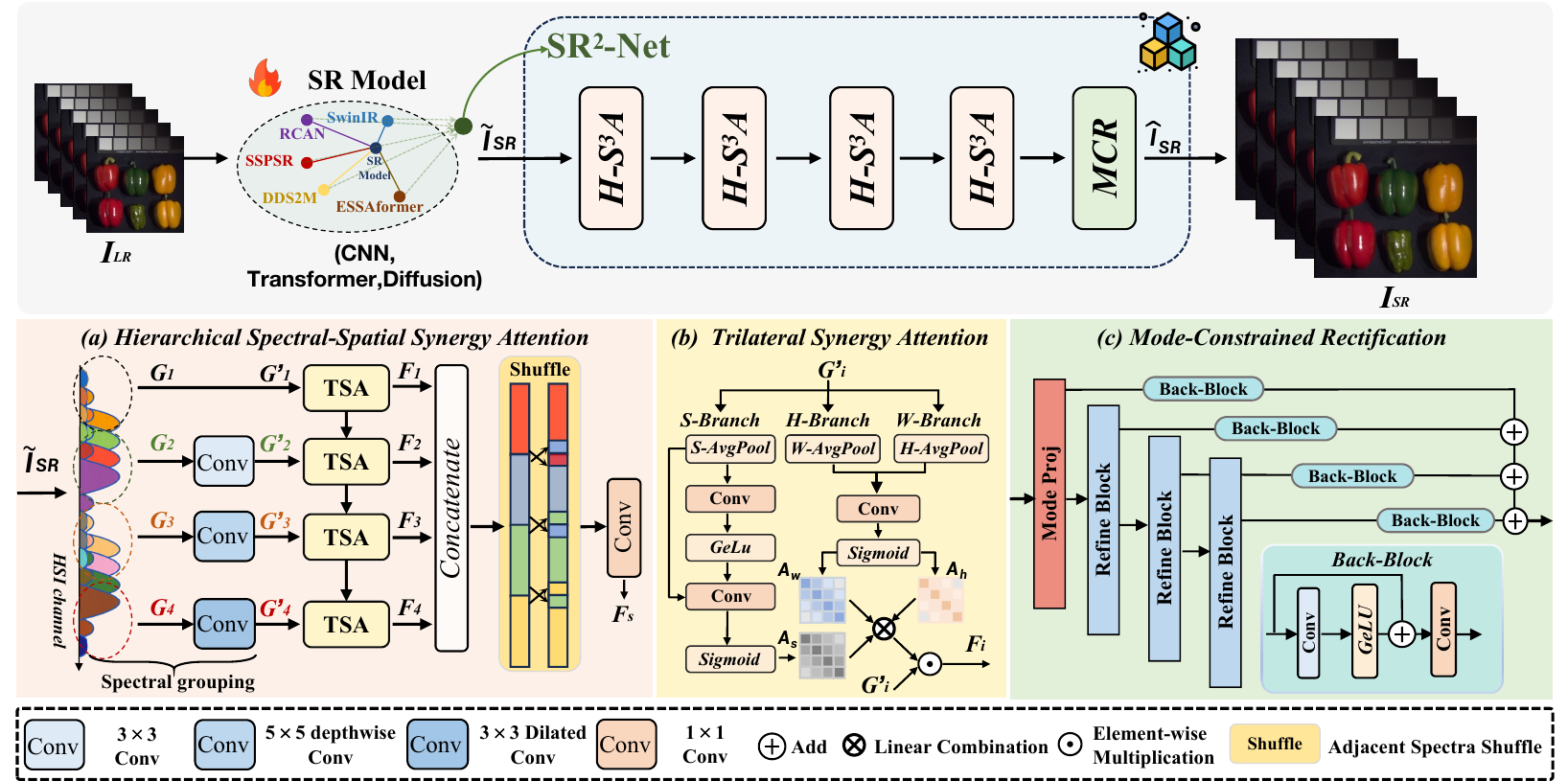}}
  \caption{Overview of~\name. Given the LR input $\textit{I}_{\mathrm{LR}}$, an SR backbone produces an initial reconstruction $\tilde{\textit{I}}_{\mathrm{SR}}$, which~\name rectifies into the output $\hat{\textit{I}}_{\mathrm{SR}}$. (a) H-S$^{3}$A enhances cross-band interaction via spectral grouping and adjacent shuffle; (b) TSA fuses spectral, height, and width views to recalibrate features; (c) MCR, drawn with $N$ stages ($N{=}1$ in our configuration), projects features onto a compact spectral subspace and aggregates the back-projected corrections.}
  \label{fig:pipeline}
\end{figure}

\name places the spectral correction after the backbone rather than inside it, as shown in Fig.~\ref{fig:pipeline}. Following an enhance-then-rectify design, an existing SR backbone first recovers a high-resolution HSI from the low-resolution input, and the rectifier then corrects that reconstruction along spectral correlations. It operates at the output resolution and uses local spatial context to guide the correction rather than performing a second super-resolution.

\paragraph{Notation and pipeline}
Given a low-resolution input $\textit{I}_{\mathrm{LR}}\in\mathbb{R}^{h\times w\times S}$ with $S$ ordered spectral bands, an SR backbone $f_{\mathrm{SR}}(\cdot)$ first produces an initial super-resolved reconstruction $\tilde{\textit{I}}_{\mathrm{SR}} = f_{\mathrm{SR}}(\textit{I}_{\mathrm{LR}})\in\mathbb{R}^{H\times W\times S}$. For an upscaling factor $\alpha$, its spatial dimensions satisfy $H=\alpha h$ and $W=\alpha w$.

The rectifier refines $\tilde{\textit{I}}_{\mathrm{SR}}$ in two stages. First, a Hierarchical Spectral-Spatial Synergy Attention~(H-S$^{3}$A) module enhances cross-band interactions via spectral grouping, trilateral synergy attention (TSA), and adjacent spectra shuffle. Next, a Mode-Constrained Rectification~(MCR) module confines the correction to a learned compact spectral subspace, suppressing residual spectral errors. Finally, a reconstruction loss and an auxiliary degradation-consistency constraint jointly train the network, tying the output back to the observed LR input.

\subsection{\texorpdfstring{H-S$^{3}$A}{H-S3A}: Cross-Band Interaction Enhancement}
The first side of the missing constraint is that backbones learn cross-band relations only implicitly. H-S$^{3}$A models them explicitly while keeping the features spatially resolved.
As shown in Fig.~\ref{fig:pipeline}(a), H-S$^{3}$A consists of three components: spectral grouping, trilateral synergy attention, \mbox{and adjacent spectra shuffle.}

\paragraph{Spectral grouping}
Since the band count $S$ need not be a multiple of four, we first zero-pad the spectra of $\tilde{{\textit{I}}}_{\mathrm{SR}}\in\mathbb{R}^{H\times W\times S}$ to $S'=4\lceil S/4\rceil$ bands; for the 31-band benchmarks this gives $S'=32$.
We then evenly split the padded spectra into four contiguous groups $\{{G}_{1},{G}_{2},{G}_{3},{G}_{4}\}$ along the band dimension, with ${G}_{i}\in\mathbb{R}^{H\times W\times S'/4}$.
The padded band sits at the tail of the last group and is discarded after the final back-projection, so the rectifier preserves the original number of bands.
The four groups use complementary per-group stems: ${G}_{1}$ keeps an identity path, while ${G}_{2}$, ${G}_{3}$, and ${G}_{4}$ are processed by a $3\times3$ convolution, a $5\times5$ depthwise convolution, and a $3\times3$ dilated convolution, respectively.
This diversifies the effective receptive field across groups at low cost, producing $\{{G}'_{1},{G}'_{2},{G}'_{3},{G}'_{4}\}$ that are fed into TSA.

\paragraph{Trilateral Synergy Attention}
We apply TSA to each group feature ${G}'_{i}\in\mathbb{R}^{H\times W\times S'/4}$ to jointly capture spectral and spatial dependencies.
TSA first summarizes ${G}'_{i}$ into three complementary descriptors by average pooling along one axis at a time:
\begin{equation}
\begin{aligned}
{A}^{h}_i&=\sigma_{g}(\mathrm{AvgPool}_{w}\,G'_i), &
{A}^{w}_i&=\sigma_{g}(\mathrm{AvgPool}_{h}\,G'_i),\\
{A}^{s}_i&=\sigma_{s}(\mathrm{AvgPool}_{s}\,G'_i), &&
\end{aligned}
\end{equation}
where the subscript of $\mathrm{AvgPool}$ is the axis averaged over.
Pooling over width gives the height descriptor ${A}^{h}_i\in\mathbb{R}^{H\times1\times S'/4}$ and pooling over height gives ${A}^{w}_i\in\mathbb{R}^{1\times W\times S'/4}$; both retain per-band coordinate context.
Pooling over the spectral axis gives ${A}^{s}_i\in\mathbb{R}^{H\times W\times1}$, which aggregates the bands into a single spatial gate.
The spectral branch $\sigma_{s}(\cdot)$ applies a convolution, a GeLU activation, a second convolution, and a sigmoid, while the height and width descriptors share a lighter gate $\sigma_{g}(\cdot)$ of a single convolution followed by a sigmoid, so every descriptor lies in $[0,1]$.
We then broadcast the three descriptors to $\mathbb{R}^{H\times W\times S'/4}$ and fuse them to recalibrate the group feature:
\begin{equation}
{F}_{i}={G}'_{i}\odot\big(\lambda_{h}{A}^{h}_i+\lambda_{w}{A}^{w}_i+\lambda_{s}{A}^{s}_i\big),
\end{equation}
where $\lambda_{h},\lambda_{w},\lambda_{s}$ are learnable scalars that weight the three broadcast descriptors and $\odot$ is element-wise multiplication.
Applying TSA to the four groups yields $\{{F}_{1},{F}_{2},{F}_{3},{F}_{4}\}$.

\paragraph{Adjacent Spectra Shuffle}

\looseness=-1 Group-wise processing keeps the cost low but leaves the groups isolated. An adjacent spectra shuffle, inspired by channel shuffle~\citep{zhang2018shufflenet}, restores exchange between them.
Specifically, we concatenate the TSA-enhanced groups along the spectral dimension, ${F}=[{F}_{1},{F}_{2},{F}_{3},{F}_{4}]\in\mathbb{R}^{H\times W\times S'}$, where $[\cdot]$ denotes concatenation.
The shuffle swaps the first channel of each group with the last channel of the preceding group, a fixed permutation designed to couple adjacent groups while preserving spectral locality.
The zero-padded channel sits at the tail of the last group and is excluded from this boundary swap.
A $1\times1$ convolution then mixes the shuffled bands, completing one H-S$^{3}$A block. Four such blocks are applied in sequence, each wrapped in a residual connection, and the output $F_{s}\in\mathbb{R}^{H\times W\times S'}$ of the last one is fed into the MCR for further rectification.

\subsection{MCR: Mode-Constrained Rectification}

\looseness=-1 Hyperspectral spectra concentrate near a low-dimensional subspace~\citep{DBLP:journals/tgrs/ChenZHZJH24}, spanned by the few principal directions that PCA exploits to compress an HSI. Mode-Constrained Rectification~(MCR) supplies the other side of the missing constraint by constraining the correction itself rather than the loss. It models the correction the way hyperspectral data are modeled, as $r=8$ learned spectral modes shared by all pixels, each pixel with its own coefficients (Sec.~\ref{sec:simple_rectifiers}). Every stage shares one bias-free back-projection $g_m(\cdot)$ whose $r$ output directions are these modes, so the aggregated correction never leaves their span and reshapes a spectrum along them rather than band by band.

As illustrated in Fig.~\ref{fig:pipeline}(c), MCR first applies a Mode Projection $P(\cdot)$ that maps the input feature $F_s \in \mathbb{R}^{H\times W\times S'}$ to a reduced dimension $r$:
\begin{equation}
m_0 = P(F_s) \in \mathbb{R}^{H\times W\times r}.
\end{equation}
MCR then applies $N$ refinement stages ($N{=}1$ in our configuration), where the $i$-th stage updates the embedding through a Refine Block $f_m(\cdot)$:
\begin{equation}
m_i = f_m(m_{i-1}), \quad i=1,2,\dots,N.
\end{equation}
For each $m_i$, a Back-Block $g_m(\cdot)$ maps the embedding back to the padded $S'$-band space; summed over all stages, these terms form a correction that is added to the backbone's reconstruction once the zero-padded band is discarded:
\begin{equation}
\hat{\textit{I}}_{\mathrm{SR}} = \tilde{\textit{I}}_{\mathrm{SR}} + \Pi_{S}\!\Big(\sum_{i=0}^{N} g_m(m_i)\Big),
\end{equation}
where $\Pi_{S}(\cdot)$ discards the zero-padded band. The $i=0$ term back-projects the un-refined embedding, and each refinement stage adds a further term to the sum.

MCR is deliberately simple: three small convolution blocks. The Mode Projection $P(\cdot)$ is a $1\times1$ convolution that reduces the channel dimension from $S'$ to $r$, followed by a GeLU activation, and the Refine Block $f_m(\cdot)$ shares this structure at dimension $r$. The Back-Block $g_m(\cdot)$ applies a $3\times3$ convolution with a GeLU activation and a residual connection at rank $r$, followed by a $1\times1$ convolution that maps $r$ back to $S'$.

\subsection{Degradation-Consistency Loss}
The degradation-consistency loss anchors the rectified output to the measurement, acting as a soft data-fidelity term for HSI-SR.
The predicted HR image $\hat{\textit{I}}_{\mathrm{SR}}$ is degraded back to the LR space by bicubic downsampling and compared with the input $\textit{I}_{\mathrm{LR}}$:
\begin{equation}
\mathcal{L}_{\mathrm{deg}} = \frac{1}{hwS}\left\| \mathcal{D}\!\left(\hat{{\textit{I}}}_{\mathrm{SR}}\right) -
{\textit{I}}_{\mathrm{LR}}
\right\|_{1},
\end{equation}
where $\mathcal{D}(\cdot)$ denotes the degradation model used to generate the LR input.
We retain the standard mean $\ell_1$ reconstruction loss $\mathcal{L}_{\mathrm{rec}}$ of the SR model, giving the total loss $\mathcal{L}_{\mathrm{total}} = \lambda_{\mathrm{rec}}\mathcal{L}_{\mathrm{rec}} + \lambda_{\mathrm{deg}}\mathcal{L}_{\mathrm{deg}}$, where $\lambda_{\mathrm{rec}}$ and $\lambda_{\mathrm{deg}}$ balance the two terms.

% Declare Table 1 early so it can occupy the top of page 6.
\begin{table}[t]
\centering
\captionsetup{skip=3pt}

\caption{In-domain results on ARAD-1K and CAVE at $\times2$, $\times4$, and $\times8$, with training and test sets drawn from disjoint splits of each dataset. Every backbone is evaluated as Base and after attaching~\name (+Ours, in bold), with one fixed rectifier configuration throughout. mPSNR and mSSIM are spatial, mSAM and CC spectral, and $\up$/$\down$ mark the better direction.}
\label{tab:main_arad_cave}
\footnotesize\renewcommand{\arraystretch}{1.1}\setlength{\tabcolsep}{2.6pt}
\begin{adjustbox}{max width=\linewidth,center}
\begin{tabular}{@{}ccc cO cO cO cO cO@{}}
\toprule
\multirow{3}{*}{Dataset} &
\multirow{3}{*}{Scale} &
\multirow{3}{*}{Metric}
& \multicolumn{4}{c}{CNN-based}
& \multicolumn{4}{c}{Transformer-based}
& \multicolumn{2}{c}{Diffusion-based}
\\
\cmidrule(lr){4-7}\cmidrule(lr){8-11}\cmidrule(l){12-13}
& &
& \multicolumn{2}{c}{RCAN} & \multicolumn{2}{c}{SSPSR}
& \multicolumn{2}{c}{SwinIR} & \multicolumn{2}{c}{ESSAformer}
& \multicolumn{2}{c}{DDS2M}
\\
\cmidrule(lr){4-5}\cmidrule(lr){6-7}\cmidrule(lr){8-9}\cmidrule(lr){10-11}\cmidrule(l){12-13}
& &
& Base & +Ours
& Base & +Ours
& Base & +Ours
& Base & +Ours
& Base & +Ours
\\
\midrule

% ===================== ARAD =====================
\multirow{12}{*}{\rotatebox[origin=c]{90}{ARAD-1K}}
& \multirow{4}{*}{$\times2$}
 & mPSNR$\up$ & 48.7390 & 50.2735 & 46.0882 & 48.5005 & 50.4662 & 51.6551 & 49.4500 & 49.9900 & 36.5792 & 37.2283 \\
 &  & mSAM$\down$ & 1.2071 & 0.7849 & 0.9906 & 0.6658 & 0.7727 & 0.4842 & 0.6090 & 0.5080 & 7.1986 & 4.7788 \\
 &  & mSSIM$\up$ & 0.9936 & 0.9964 & 0.9893 & 0.9920 & 0.9972 & 0.9975 & 0.9937 & 0.9942 & 0.9089 & 0.9277 \\
 &  & CC$\up$ & 0.9978 & 0.9990 & 0.9960 & 0.9981 & 0.9990 & 0.9993 & 0.9986 & 0.9987 & 0.9723 & 0.9799 \\
\cmidrule(l){2-13}

& \multirow{4}{*}{$\times4$}
 & mPSNR$\up$ & 38.3786 & 40.0972 & 38.8792 & 40.1004 & 39.5717 & 40.9720 & 39.9537 & 40.7200 & 30.8476 & 33.2613 \\
 &  & mSAM$\down$ & 1.9009 & 1.3010 & 1.6076 & 1.3128 & 1.3950 & 1.2819 & 1.3670 & 1.2300 & 7.0404 & 5.8757 \\
 &  & mSSIM$\up$ & 0.9470 & 0.9719 & 0.9463 & 0.9543 & 0.9709 & 0.9752 & 0.9592 & 0.9599 & 0.7984 & 0.8671 \\
 &  & CC$\up$ & 0.9812 & 0.9916 & 0.9858 & 0.9891 & 0.9902 & 0.9924 & 0.9903 & 0.9904 & 0.9421 & 0.9549 \\
\cmidrule(l){2-13}

& \multirow{4}{*}{$\times8$}
 & mPSNR$\up$ & 33.3062 & 34.3745 & 34.1149 & 34.9975 & 33.4185 & 34.8318 & 34.6531 & 35.0546 & 27.7449 & 29.5794 \\
 &  & mSAM$\down$ & 3.0518 & 2.1900 & 2.4223 & 2.0352 & 2.4682 & 2.1164 & 2.3531 & 2.0140 & 11.5589 & 9.2067 \\
 &  & mSSIM$\up$ & 0.8891 & 0.9029 & 0.8724 & 0.8850 & 0.8869 & 0.9103 & 0.8882 & 0.8889 & 0.6060 & 0.7993 \\
 &  & CC$\up$ & 0.9628 & 0.9693 & 0.9622 & 0.9681 & 0.9551 & 0.9720 & 0.9691 & 0.9698 & 0.8861 & 0.9019 \\

\midrule

% ===================== CAVE =====================
\multirow{12}{*}{\rotatebox[origin=c]{90}{CAVE}}
& \multirow{4}{*}{$\times2$}
 & mPSNR$\up$ & 41.0944 & 43.4074 & 39.4003 & 40.4833 & 40.3123 & 41.4861 & 39.4700 & 40.7900 & 35.5928 & 36.5567 \\
 &  & mSAM$\down$ & 5.7619 & 4.4010 & 4.2518 & 3.7893 & 0.8214 & 0.6275 & 4.0820 & 3.3450 & 11.0914 & 8.1802 \\
 &  & mSSIM$\up$ & 0.9838 & 0.9897 & 0.9711 & 0.9830 & 0.9754 & 0.9877 & 0.9704 & 0.9747 & 0.9053 & 0.9357 \\
 &  & CC$\up$ & 0.9946 & 0.9965 & 0.9907 & 0.9943 & 0.9807 & 0.9905 & 0.9950 & 0.9960 & 0.9882 & 0.9908 \\
\cmidrule(l){2-13}

& \multirow{4}{*}{$\times4$}
 & mPSNR$\up$ & 35.9929 & 37.4461 & 35.2386 & 35.8929 & 37.2675 & 38.1303 & 34.1100 & 35.3300 & 29.8048 & 31.4389 \\
 &  & mSAM$\down$ & 7.6445 & 5.4742 & 4.8735 & 4.4121 & 0.9015 & 0.7083 & 5.6070 & 4.5410 & 13.7359 & 10.0409 \\
 &  & mSSIM$\up$ & 0.9536 & 0.9643 & 0.9384 & 0.9427 & 0.9492 & 0.9577 & 0.9267 & 0.9367 & 0.7493 & 0.8875 \\
 &  & CC$\up$ & 0.9825 & 0.9856 & 0.9795 & 0.9807 & 0.9315 & 0.9478 & 0.9846 & 0.9875 & 0.9699 & 0.9766 \\
\cmidrule(l){2-13}

& \multirow{4}{*}{$\times8$}
 & mPSNR$\up$ & 32.0839 & 33.2764 & 31.5901 & 32.0282 & 32.8788 & 34.2636 & 30.0600 & 30.8700 & 27.1577 & 28.1137 \\
 &  & mSAM$\down$ & 10.1641 & 6.8252 & 7.6479 & 6.1349 & 1.5060 & 0.8553 & 7.8170 & 6.0120 & 16.6298 & 13.7381 \\
 &  & mSSIM$\up$ & 0.8923 & 0.9104 & 0.8781 & 0.8828 & 0.9125 & 0.9321 & 0.8530 & 0.8664 & 0.6298 & 0.7948 \\
 &  & CC$\up$ & 0.9616 & 0.9654 & 0.9577 & 0.9599 & 0.9185 & 0.9233 & 0.9637 & 0.9683 & 0.9186 & 0.9389 \\

\bottomrule

\end{tabular}
\end{adjustbox}

\end{table}

\section{Experiments}

\subsection{Experimental Settings}
\label{sec:settings}
\paragraph{Datasets and degradation}
We evaluate on three widely used 31-band hyperspectral datasets: ARAD-1K~\citep{Arad_2022_CVPR_recovery} with the standard $900/50$ train/test split, CAVE~\citep{park2007multispectral} with $16/16$ scenes, and $18$ ICVL~\citep{arad2016sparse} HSIs reserved for cross-domain evaluation only.
LR inputs are synthesized by bicubic downsampling at each scale factor, and robustness to test-time degradation mismatch is examined separately without changing this training setup.

\paragraph{Evaluation metrics}
We measure spatial reconstruction with mean PSNR~(mPSNR) and mean SSIM~(mSSIM) \citep{wang2004image}, and spectral fidelity with mean spectral angle mapper~(mSAM, in degrees)~\citep{kruse1993spectral} and the mean spectral correlation coefficient~(CC)~\citep{loncan2015hyperspectral}.
Most tables report mPSNR, mSSIM, and mSAM, with formal definitions in Appendix~\ref{sec:supp_metrics}.

\paragraph{Compared methods}
To verify model-agnostic generality, we attach~\name to five representative baseline models:
CNN-based RCAN~\citep{zhang2018image} and SSPSR~\citep{jiang2020learning},
Transformer-based SwinIR~\citep{liang2021swinir} and ESSAformer~\citep{zhang2023essaformer},
and diffusion-based DDS2M~\citep{miao2023dds2m}.

For each SR backbone $\mathcal{B}$, we evaluate the baseline $\mathcal{B}$ and the rectified version $\mathcal{B}+\mathrm{SR}^{2}\text{-Net}$ under the same settings.
\name is therefore evaluated as an architecture-preserving refinement. Each of the four feed-forward backbones is trained jointly with the rectifier (frozen-backbone results are in Sec.~\ref{sec:simple_rectifiers}). Their Base models are trained with the reconstruction loss only, whereas the degradation-consistency constraint applies to the rectified models. The diffusion-based DDS2M instead reconstructs each scene by self-supervised per-image inference; the rectifier is trained on its outputs for the training scenes under the same losses and evaluated on the held-out test scenes.

\paragraph{Implementation details}
\looseness=-1 All networks are implemented in PyTorch on an NVIDIA RTX 3090 GPU and trained using AdamW, with the learning rate decayed from $1\times10^{-4}$ to $1\times10^{-5}$ by cosine annealing over 1000 epochs.
We optimize the mean $\ell_1$ reconstruction loss together with the degradation-consistency constraint, using $\lambda_{\mathrm{rec}}=1.0$ and $\lambda_{\mathrm{deg}}=0.1$, and~\name uses 4 spectral groups, $4$ H-S$^{3}$A blocks, and $1$ MCR block with bottleneck rank $r=8$.
Full hyperparameters are listed in Appendix~\ref{sec:supp_config}.

\subsection{Quantitative Results and Analysis}

\paragraph{In-domain performance}
Training and testing within each of ARAD-1K and CAVE, Tab.~\ref{tab:main_arad_cave} covers $\times2$, $\times4$, and $\times8$. Spectral fidelity improves in all thirty configurations, with mSAM falling by $1.1203$ on average and by at least $0.1010$. The gain is not bought with spatial quality: mPSNR rises by between $0.4015$ and $2.4137$~dB, and mSSIM and CC \mbox{never decrease.}

\looseness=-1 How much the rectifier corrects follows how much spectral error the backbone leaves. The largest gains occur for the diffusion-based DDS2M, whose mSAM drops by $2.4198$ on ARAD-1K at $\times2$ and by $3.6950$ on CAVE at $\times4$, whereas the spectrally strong ESSAformer leaves the least to correct ($0.1010$ on ARAD-1K at $\times2$). Over the thirty configurations the reduction tracks the Base mSAM with a correlation of $0.93$, and one fixed configuration removes $22.6\%$ of that error on average, with per-backbone means between $16.9\%$ and $29.9\%$. The pattern persists as the task gets harder, since the reduction at $\times8$ exceeds the one at $\times2$ for all eight feed-forward combinations of backbone and dataset.
% Cross-domain table: the two scales sit side by side to span the text width.
\begin{table}[t]
\centering
\captionsetup{skip=3pt}
\caption{Cross-domain results, with every backbone trained on ARAD-1K and tested on CAVE and ICVL without fine-tuning. Each backbone is evaluated as Base and after attaching~\name (+Ours, in bold) under the identical protocol.}
\label{tab:cross_domain}
\footnotesize\renewcommand{\arraystretch}{1.1}\setlength{\tabcolsep}{3.2pt}
\begin{adjustbox}{max width=\linewidth,center}
\begin{tabular}{@{}ll cccc cccc@{}}
\toprule
\multirow{3}{*}{Method} & \multirow{3}{*}{Variant} & \multicolumn{4}{c}{$\times2$} & \multicolumn{4}{c}{$\times4$} \\
\cmidrule(lr){3-6}\cmidrule(l){7-10}
 & & \multicolumn{2}{c}{CAVE} & \multicolumn{2}{c}{ICVL} & \multicolumn{2}{c}{CAVE} & \multicolumn{2}{c}{ICVL} \\
\cmidrule(lr){3-4}\cmidrule(lr){5-6}\cmidrule(lr){7-8}\cmidrule(l){9-10}
 & & mPSNR\Up & mSAM\Down & mPSNR\Up & mSAM\Down & mPSNR\Up & mSAM\Down & mPSNR\Up & mSAM\Down \\
\midrule
\multirow{2}{*}{RCAN} & Base & 42.7269 & 5.0004 & 49.6448 & 0.7389 & 35.6214 & 6.2153 & 39.9389 & 1.0969 \\
 & \cellcolor{oursrow}+Ours & \cellcolor{oursrow}\textbf{44.2866} & \cellcolor{oursrow}\textbf{4.0978} & \cellcolor{oursrow}\textbf{51.8506} & \cellcolor{oursrow}\textbf{0.4682} & \cellcolor{oursrow}\textbf{36.3326} & \cellcolor{oursrow}\textbf{5.3594} & \cellcolor{oursrow}\textbf{40.9148} & \cellcolor{oursrow}\textbf{0.7897} \\
\midrule
\multirow{2}{*}{SSPSR} & Base & 41.4071 & 4.0419 & 48.4758 & 0.8140 & 36.7963 & 4.9284 & 43.9812 & 1.0271 \\
 & \cellcolor{oursrow}+Ours & \cellcolor{oursrow}\textbf{43.0076} & \cellcolor{oursrow}\textbf{2.9243} & \cellcolor{oursrow}\textbf{50.7217} & \cellcolor{oursrow}\textbf{0.5264} & \cellcolor{oursrow}\textbf{37.5721} & \cellcolor{oursrow}\textbf{3.9189} & \cellcolor{oursrow}\textbf{45.1945} & \cellcolor{oursrow}\textbf{0.6698} \\
\midrule
\multirow{2}{*}{SwinIR} & Base & 40.7157 & 4.4366 & 47.9367 & 0.8260 & 34.7725 & 4.8551 & 41.0743 & 1.0552 \\
 & \cellcolor{oursrow}+Ours & \cellcolor{oursrow}\textbf{42.3415} & \cellcolor{oursrow}\textbf{2.6231} & \cellcolor{oursrow}\textbf{50.0704} & \cellcolor{oursrow}\textbf{0.4709} & \cellcolor{oursrow}\textbf{35.7256} & \cellcolor{oursrow}\textbf{3.8360} & \cellcolor{oursrow}\textbf{42.6530} & \cellcolor{oursrow}\textbf{0.6682} \\
\midrule
\multirow{2}{*}{ESSAformer} & Base & 42.1763 & 3.3575 & 49.7895 & 0.6404 & 37.0305 & 4.8258 & 44.3869 & 0.9349 \\
 & \cellcolor{oursrow}+Ours & \cellcolor{oursrow}\textbf{43.4978} & \cellcolor{oursrow}\textbf{2.5611} & \cellcolor{oursrow}\textbf{51.3981} & \cellcolor{oursrow}\textbf{0.4625} & \cellcolor{oursrow}\textbf{37.6350} & \cellcolor{oursrow}\textbf{3.5556} & \cellcolor{oursrow}\textbf{45.7012} & \cellcolor{oursrow}\textbf{0.6079} \\
\bottomrule
\end{tabular}
\end{adjustbox}
\end{table}

\paragraph{Cross-domain generalization}
Trained on the 900 ARAD-1K training images and tested on all CAVE scenes and all ICVL HSIs without fine-tuning, Tab.~\ref{tab:cross_domain} covers $\times2$ and $\times4$. Models generalize better under domain shift after attaching~\name, with mSAM reductions in every reported configuration while preserving spatial metrics.
For instance, SwinIR on CAVE at $\times2$ improves mPSNR from $40.7157$ to $42.3415$ and reduces mSAM from $4.4366$ to $2.6231$.
DDS2M is reported in the in-domain setting only, as its diffusion inference is far costlier than the other backbones.
Full results including mSSIM are in Appendix~\ref{sec:supp_cross}.

\paragraph{Overhead and robustness}
\name adds a backbone-independent overhead of $+0.048$M parameters in the 31-band setting, and its FLOPs overhead depends only on the output spatial size, amounting to $+7.7\%$ at $\times4$ and $+22.9\%$ at $\times8$ for the lightest backbone SwinIR and proportionally less for heavier ones.
It also remains effective under mild test-time mismatch. Keeping SwinIR trained under standard bicubic degradation, mSAM improves over Base in all nine test-time settings and scales (with the B1 blur at $\times4$ it drops from $1.4760$ to $1.2709$), with full results in Appendix~\ref{sec:supp_robust}.

\begin{figure}[t]
\centering
\includegraphics[width=0.94\textwidth]{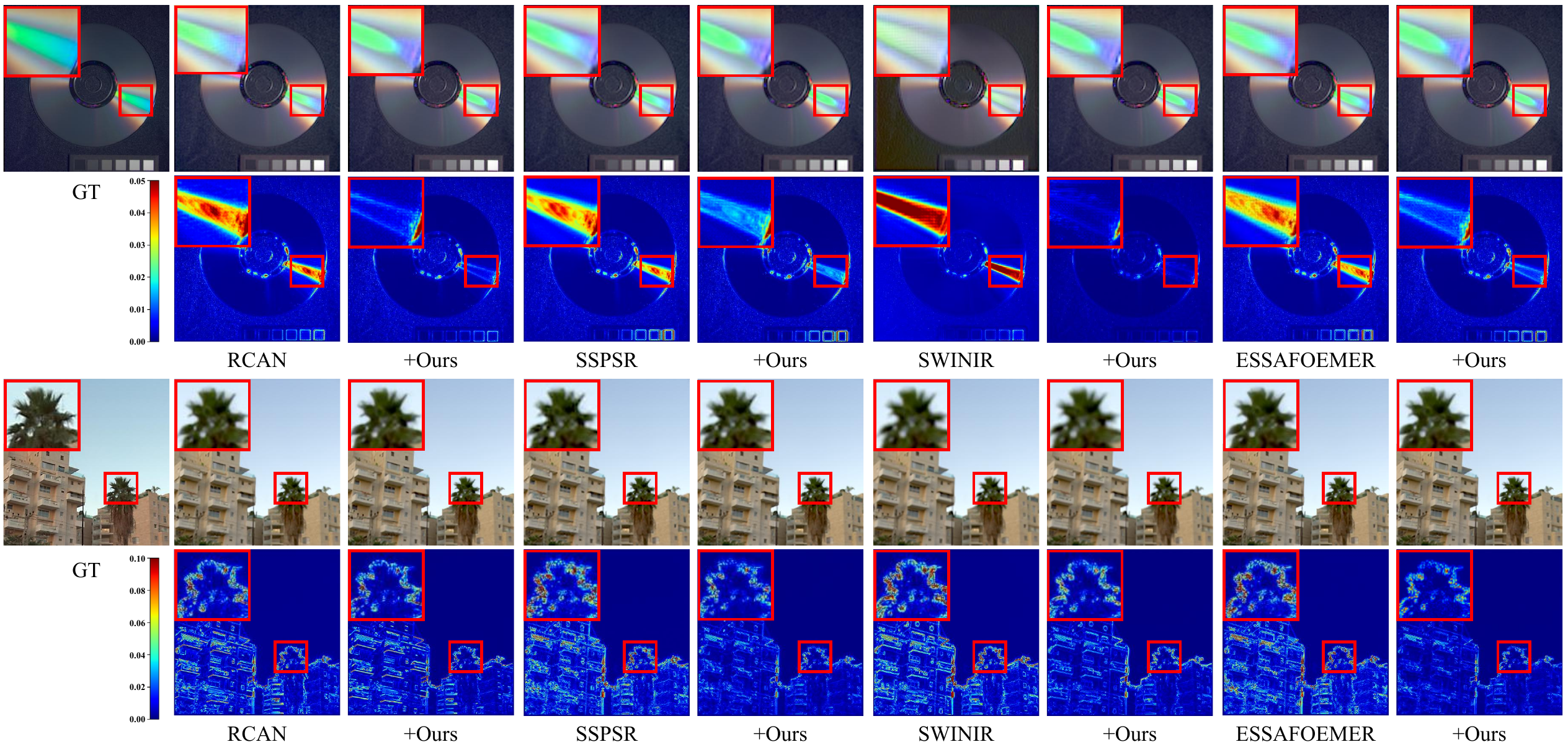}
\caption{Error maps at $\times4$ (in-domain, CAVE and ARAD-1K). Pseudo-RGB uses bands $25$/$15$/$5$ as R--G--B, and band-averaged absolute error maps (darker is better) pair each +Ours result with its Base. Red boxes mark where~\name sharpens edges and fine structures; the cross-domain ICVL comparison and more galleries are in Appendix~\ref{sec:supp_errmap}.}
\label{fig:error_map}
\end{figure}

\begin{figure}[t]
\centering
\includegraphics[width=0.96\textwidth]{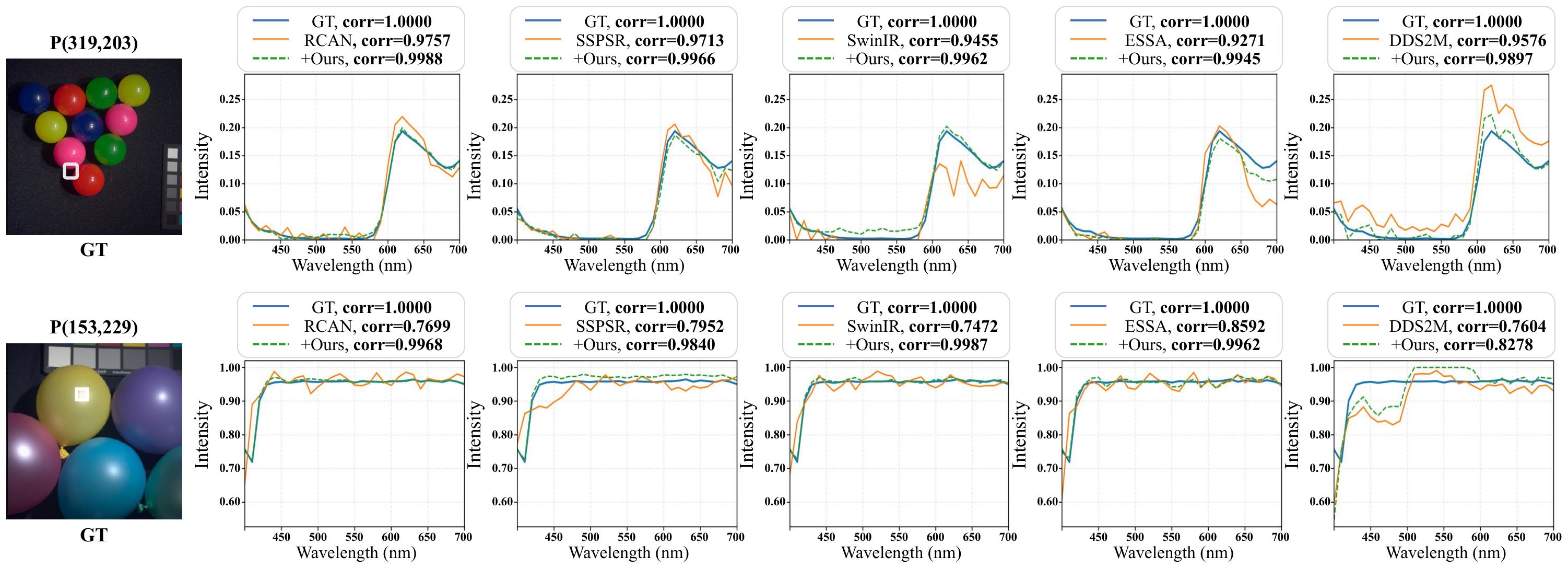}
\caption{Per-pixel spectra at $\times4$ on CAVE across CNN-, Transformer-, and diffusion-based backbones. For two representative pixels (marked in the pseudo-RGB images), the baseline spectra oscillate away from the ground truth while~\name follows it; additional scenes and \mbox{datasets are in Appendix~\ref{sec:supp_spectral}.}}
\label{fig:spectral_curve}
\end{figure}

\subsection{Visualization}
In Fig.~\ref{fig:error_map}, baseline SR models produce localized errors near object boundaries and thin structures, visible as halos.
After attaching~\name these errors are visibly reduced while fine textures and sharp edges are preserved; per-scene galleries are in Appendix~\ref{sec:supp_errmap}.

Fig.~\ref{fig:spectral_curve} visualizes per-pixel spectra under $\times4$ upsampling at CAVE locations across CNN-, Transformer-, and diffusion-based backbones.
The baseline spectra oscillate away from the ground-truth trend, while~\name yields smoother profiles that follow it on these pixels.
Beyond these examples, the effect is dataset-wide. On the 50 ARAD-1K test scenes \name lowers the residual spectral error on all 50 and improves per-scene mSAM on all 50 (one-sided Wilcoxon signed-rank, $p<10^{-6}$), and the residual analysis is in Appendix~\ref{sec:supp_significance}.

% Tables 3-5 share one float: the two ablations stacked on the left, the rectifier comparison on the right.
% Ragged-right captions avoid the stretched lines that justification forces in the narrow columns.
\begin{table}[t]
\centering
\captionsetup{skip=3pt,justification=raggedright,singlelinecheck=false}
\begin{minipage}[t]{0.5\linewidth}
\centering
\caption{Component ablation on ARAD-1K at $\times4$ with SwinIR. Best in \textbf{bold}, second \underline{underlined}.}
\label{tab:ablation_comp}
\footnotesize\renewcommand{\arraystretch}{1.1}\setlength{\tabcolsep}{3.5pt}
\begin{adjustbox}{max width=\linewidth,center}
\begin{tabular}{@{}ccccc@{}}
\toprule
\multicolumn{2}{c}{Method}  & \multicolumn{3}{c}{Metric} \\
\cmidrule(lr){1-2}\cmidrule(lr){3-5}
H-S$^{3}$A & MCR
& mPSNR$\uparrow$ & mSSIM$\uparrow$ & mSAM$\downarrow$ \\
\midrule
$\times$ & $\times$ & 39.5717 & 0.9709 & 1.3950 \\
\checkmark & $\times$ & \underline{40.7059} & \underline{0.9735} & 1.3476 \\
$\times$ & \checkmark & 40.2550 & 0.9717 & \underline{1.3173} \\
\rowcolor{oursrow}\checkmark & \checkmark & \textbf{40.9720} & \textbf{0.9752} & \textbf{1.2819} \\
\bottomrule
\end{tabular}
\end{adjustbox}
% Match the combined left tables to the bottom rule of Table 5.
\par\vspace{7.4pt}
\caption{Loss-weight ablation on ARAD-1K at $\times4$ with SwinIR. Best in \textbf{bold}, second \underline{underlined}.}
\label{tab:ablation_loss}
\footnotesize\renewcommand{\arraystretch}{1.1}\setlength{\tabcolsep}{3.5pt}
\begin{adjustbox}{max width=\linewidth,center}
\begin{tabular}{@{}ccccc@{}}
\toprule
$\lambda_{\mathrm{rec}}$ & $\lambda_{\mathrm{deg}}$
& mPSNR$\uparrow$ & mSSIM$\uparrow$ & mSAM$\downarrow$ \\
\midrule
1.0 & 0.0 & 40.6920 & 0.9740 & 1.3262 \\
\rowcolor{oursrow}1.0 & 0.1 & \textbf{40.9720} & \textbf{0.9752} & \textbf{1.2819} \\
1.0 & 0.2 & \underline{40.9225} & \underline{0.9751} & \underline{1.2833} \\
1.0 & 0.5 & 40.7184 & 0.9745 & 1.3174 \\
1.0 & 1.0 & 40.6217 & 0.9739 & 1.3305 \\
\bottomrule
\end{tabular}
\end{adjustbox}
\end{minipage}\hfill
\begin{minipage}[t]{0.465\linewidth}
\centering
\caption{Comparison to simpler rectifiers (ARAD-1K, $\times4$, SwinIR). SG/PCA/IBP are fixed post-hoc rectifiers; ``$+\mathcal{L}_{\mathrm{deg}}$ only'' adds the loss to the Base with no rectifier, and ``Generic head'' is a same-budget ($+0.048$M) residual head. +MCR and~\name are the jointly trained MCR-only and full models. Best in \textbf{bold}, second \underline{underlined}.}
\label{tab:simple_rectifiers}
% 1.25 lines this table's last row up with that of Table 4 on the left
\footnotesize\renewcommand{\arraystretch}{1.25}\setlength{\tabcolsep}{2.5pt}
\begin{adjustbox}{max width=\linewidth,center}
\begin{tabular}{@{}lcccc@{}}
\toprule
Method & mPSNR$\uparrow$ & mSAM$\downarrow$ & mSSIM$\uparrow$ & CC$\uparrow$ \\
\midrule
Base (SwinIR) & 39.5717 & 1.3950 & 0.9709 & 0.9902 \\
\midrule
\;\;+ SG      & 38.7834 & 2.4512 & 0.9698 & 0.9889 \\
\;\;+ PCA     & 39.5729 & 1.4028 & 0.9709 & 0.9902 \\
\;\;+ IBP     & 39.5782 & 1.3693 & 0.9709 & 0.9902 \\
\midrule
\;\;+ $\mathcal{L}_{\mathrm{deg}}$ only        & 39.7097 & 1.3692 & 0.9711 & 0.9905 \\
\;\;+ Generic head        & 39.9815 & 1.3648 & 0.9712 & 0.9907 \\
\midrule
\;\;+ MCR     & \underline{40.2550} & \underline{1.3173} & \underline{0.9717} & \underline{0.9912} \\
\rowcolor{oursrow}\;\;+~\name   & \textbf{40.9720} & \textbf{1.2819} & \textbf{0.9752} & \textbf{0.9924} \\
\bottomrule
\end{tabular}
\end{adjustbox}
\end{minipage}
\vspace{-10pt}
\end{table}

\subsection{Ablation Study}
\paragraph{Component ablation}
\looseness=-1 All ablations use ARAD-1K at $\times4$ with SwinIR. Tab.~\ref{tab:ablation_comp} evaluates the two-stage design of~\name.
Each stage moves the axis it was designed for. In isolation, H-S$^{3}$A contributes most of the spatial gain (mPSNR $40.7059$ against $39.5717$) but little of the spectral one, whereas MCR gives the larger spectral gain (mSAM $1.3173$ against $1.3950$) at roughly half the spatial improvement. The two are complementary rather than additive, and the full model is the best on every metric, reaching $40.9720$ mPSNR and $1.2819$ mSAM. Removing only the adjacent spectra shuffle, with the $1\times1$ mixing kept, raises mSAM to $1.3102$ (Appendix~\ref{sec:supp_struct}).

\paragraph{Degradation-consistency loss}
\looseness=-1 Tab.~\ref{tab:ablation_loss} varies $\lambda_{\mathrm{deg}}$. Removing the constraint ($\lambda_{\mathrm{deg}}=0$) raises mSAM from $1.2819$ to $1.3262$, still below the Base ($1.3950$). We select $\lambda_{\mathrm{deg}}=0.1$ once on this configuration and fix it everywhere else; $\lambda_{\mathrm{deg}}=0.2$ is within one seed standard deviation of it.

\vspace{-8pt}
\subsection{Isolating the Source of the Improvement}
\label{sec:simple_rectifiers}
We ask whether the gain comes from a cheaper alternative (Tab.~\ref{tab:simple_rectifiers}).
Among the fixed post-hoc rectifiers, Savitzky--Golay smoothing (SG) degrades every metric (mSAM $2.4512$), while iterative back-projection (IBP) and a PCA projection of the same rank as MCR barely change mSAM ($1.3693$ and $1.4028$).
The degradation-consistency loss alone ($1.3692$) and a same-budget generic residual head ($1.3648$) stay far above MCR ($1.3173$) and the full~\name ($1.2819$); the generic head shares the full model's $+0.048$M budget, whereas MCR alone uses less.
Nor is the gain a capacity effect: narrowing the bottleneck from $r=31$ to $r=8$ lowers mSAM from $1.331$ to $1.2819$ while removing parameters, and $r=1$ is worse again ($1.349$), so the rank acts as a two-sided constraint. The rectifier also works with the four backbones frozen at their Base weights: mSAM falls in all sixteen ARAD-1K and CAVE configurations at $\times2$ and $\times4$, recovering $76\%$ of the jointly trained gain on average (Tab.~\ref{tab:frozen_backbone}). The ranking from Base through the cheap controls to MCR and the full model is stable across three seeds, and the full model also leads the fixed rectifiers and MCR at $\times2$ and $\times8$; implementation, seed stability, and the structural sweeps are in Appendices~\ref{sec:supp_simple} and~\ref{sec:supp_struct}.

\vspace{-8pt}
\section{Conclusion}
\looseness=-1 We presented~\name, a model-agnostic spectral rectifier for hyperspectral image super-resolution.
Following an enhance-then-rectify design, it attaches to an SR backbone's output without modifying its internal architecture: H-S$^{3}$A strengthens cross-band interactions, MCR confines the correction to a compact spectral subspace, and a degradation-consistency constraint ties the output to the observed LR input.
Across five backbones spanning CNN, Transformer, and diffusion families, it improves spectral fidelity in every reported setting while preserving or improving spatial quality, in domain, under cross-domain transfer (four of five backbones), and under mild test-time mismatch, at a constant $+0.048$M overhead.

% End-of-paper statements.
\section*{AI Use Statement}
\looseness=-1 We have not used generative AI tools to develop the method or its conceptual framework, to formulate hypotheses or claims, to design the experiments, to implement the method, to generate or process data, to translate, or to interpret the results.
We used generative AI tools only to edit and rephrase the text of this paper for readability.
We have reviewed all AI-assisted edits, and we take responsibility for the final content of this work, including text, claims or artifacts produced with the aid of generative AI.

\section*{Ethics Statement}
This work uses only the public ARAD-1K, CAVE, and ICVL hyperspectral benchmarks under their original terms of use and involves no new data collection or human subjects. We are not aware of ethical concerns specific to this work.

\section*{Reproducibility Statement}
All experiments use the public ARAD-1K, CAVE, and ICVL datasets with the splits and bicubic degradation described in Section~\ref{sec:settings}.
Appendix~\ref{sec:supp_metrics} defines every metric and the border cropping applied before evaluation.
Appendix~\ref{sec:supp_config} lists the training hyperparameters and the full rectifier configuration (Tables~\ref{tab:supp_hparams} and~\ref{tab:supp_arch}), shared across backbones and scales except for the scale-dependent patch size.
Appendix~\ref{sec:supp_deg} specifies the test-time degradation protocol, including the per-image seeding of the added noise, and Appendix~\ref{sec:supp_simple} details the fixed and learned control rectifiers and reports seed stability over three seeds.
The code will be released upon publication.

% ---------------- Bibliography ----------------
\bibliography{references}
\bibliographystyle{iclr2027_conference}

% ---------------- Appendix ----------------
% Appendix floats and equations are numbered A1, A2, ... so that they cannot be
% confused with the main-text numbering; the \theH* variants keep hyperref's
% link targets unique after the counters restart.
\clearpage
\appendix
\renewcommand{\thetable}{A\arabic{table}}
\renewcommand{\thefigure}{A\arabic{figure}}
\renewcommand{\theequation}{A\arabic{equation}}
\renewcommand{\theHtable}{A\arabic{table}}
\renewcommand{\theHfigure}{A\arabic{figure}}
\renewcommand{\theHequation}{A\arabic{equation}}
\setcounter{table}{0}
\setcounter{figure}{0}
\setcounter{equation}{0}
% Technical appendix, input by main.tex after \appendix and the references.
% Appendix tables typeset +Ours cells in bold on the light-blue background and Gain cells plain.
\renewcommand{\ourscell}[1]{\cellcolor{oursrow}\textbf{#1}}
\newcolumntype{P}{>{\columncolor{oursrow}}c}  % shaded column for the frozen-backbone +Ours results
\renewcommand{\gainpos}[1]{#1}

% Float pages start at the top margin instead of being vertically centered.
\makeatletter
\setlength{\@fptop}{0pt}
\makeatother

% Full-width gallery block: three stacked panels with one figure caption.
\newcommand{\ErrMapGalleryBlock}[6]{%
\noindent\begin{minipage}{\textwidth}
\centering
\captionsetup{skip=3pt,hypcap=false}
\includegraphics[width=0.98\linewidth,height=#6\textheight,keepaspectratio]{#2}\\[2pt]
\includegraphics[width=0.98\linewidth,height=#6\textheight,keepaspectratio]{#3}\\[2pt]
\includegraphics[width=0.98\linewidth,height=#6\textheight,keepaspectratio]{#4}
\captionof{figure}{#1}
\label{#5}
\end{minipage}%
}

\section*{Appendix}
\looseness=-1 This appendix provides the formal metric definitions, full configurations and protocols, complete quantitative results, and additional qualitative galleries complementing the main paper. The quantitative sections (Sections~\ref{sec:supp_metrics}--\ref{sec:supp_significance}) are self-contained and precede the qualitative galleries (Sections~\ref{sec:supp_errmap} and~\ref{sec:supp_spectral}).

\section{Evaluation Metrics}
\label{sec:supp_metrics}
Let $\textit{X}\in\mathbb{R}^{H\times W\times S}$ denote the ground-truth HSI and $\hat{\textit{X}}$ its reconstruction, with $S$ spectral bands. We use two spatial-fidelity metrics (mPSNR and mSSIM) and two spectral-fidelity metrics (mSAM and CC). Spatial metrics are averaged over the $S$ bands, whereas spectral metrics are averaged over pixels and images. Unless noted otherwise, LR inputs are produced by bicubic downsampling of the HR cube, all cubes are scaled to $[0,1]$, and each metric is computed after cropping a border of $s$ pixels ($s$ the scale factor) to remove boundary artifacts.

\paragraph{Mean PSNR (mPSNR)}
For each band $b$, the peak signal-to-noise ratio is
\begin{equation}
\mathrm{PSNR}_b = 10\log_{10}\!\left(\frac{\mathrm{MAX}^{2}}{\frac{1}{HW}\sum_{p}\big(\textit{X}_{b,p}-\hat{\textit{X}}_{b,p}\big)^{2}}\right),
\end{equation}
where the sum runs over the $HW$ spatial positions $p$ and $\mathrm{MAX}$ is the dynamic range of the data. mPSNR averages $\mathrm{PSNR}_b$ over all bands; higher is better.

\paragraph{Mean SSIM (mSSIM)}
For a reference band $x$ and its reconstruction $y$, the structural similarity is
\begin{equation}
\mathrm{SSIM}(x,y)=\frac{(2\mu_x\mu_y+C_1)(2\sigma_{xy}+C_2)}{(\mu_x^{2}+\mu_y^{2}+C_1)(\sigma_x^{2}+\sigma_y^{2}+C_2)},
\end{equation}
where $\mu$, $\sigma^{2}$, and $\sigma_{xy}$ are the local means, variances, and covariance, and $C_1,C_2$ are stabilizing constants. mSSIM averages the per-band SSIM; higher is better.

\paragraph{Mean SAM (mSAM)}
The spectral angle mapper measures the angle between the reference and reconstructed spectra at each pixel $p$,
\begin{equation}
\mathrm{SAM}_p=\arccos\!\left(\frac{\langle \textit{x}_p,\hat{\textit{x}}_p\rangle}{\lVert \textit{x}_p\rVert_2\,\lVert \hat{\textit{x}}_p\rVert_2}\right),
\end{equation}
where $\textit{x}_p,\hat{\textit{x}}_p\in\mathbb{R}^{S}$ are the spectra at pixel $p$. mSAM averages $\mathrm{SAM}_p$ over all pixels and images and is reported in degrees; lower is better. Because it compares the direction of the spectral vector rather than its magnitude, mSAM is the primary indicator of spectral fidelity in this paper.

\paragraph{Spectral correlation coefficient (CC)}
CC is the Pearson correlation between the reference and reconstructed spectra at each pixel,
\begin{equation}
\mathrm{CC}_p=\frac{\sum_{s}(\textit{x}_{p,s}-\bar{\textit{x}}_p)(\hat{\textit{x}}_{p,s}-\bar{\hat{\textit{x}}}_p)}{\sqrt{\sum_{s}(\textit{x}_{p,s}-\bar{\textit{x}}_p)^{2}}\,\sqrt{\sum_{s}(\hat{\textit{x}}_{p,s}-\bar{\hat{\textit{x}}}_p)^{2}}},
\end{equation}
\looseness=-1 where $\bar{\textit{x}}_p$ and $\bar{\hat{\textit{x}}}_p$ are the per-pixel spectral means. CC is averaged over pixels and images; higher is better.

\section{Network Configuration and Training Details}
\label{sec:supp_config}
Table~\ref{tab:supp_hparams} summarizes the training hyperparameters, and Table~\ref{tab:supp_arch} lists the configuration of the two rectification stages. Unless stated otherwise, all settings are shared across backbones and scales; only the patch size depends on the scale factor. These values match those used to produce every result reported in the main paper.

\begin{table}[!htbp]
\centering
\footnotesize
\setlength{\tabcolsep}{4pt}
\renewcommand{\arraystretch}{1.2}

\caption{Training hyperparameters, shared across all backbones.}
\label{tab:supp_hparams}
\begin{tabular}{@{}ll@{}}
\toprule
Setting & Value \\
\midrule
Framework & PyTorch \\
Hardware & NVIDIA RTX 3090 GPU \\
Optimizer & AdamW ($\beta_1=0.9$, $\beta_2=0.99$) \\
Learning rate & $1\times10^{-4}\rightarrow1\times10^{-5}$ (cosine) \\
Epochs & 1000 \\
Batch size & 8 \\
LR patch ($\times2$/$\times4$/$\times8$) & $64^2$ / $32^2$ / $32^2$ \\
HR patch ($\times2$/$\times4$/$\times8$) & $128^2$ / $128^2$ / $256^2$ \\
Reconstruction loss & mean $\ell_1$, $\lambda_{\mathrm{rec}}=1.0$ \\
Degradation loss & mean $\ell_1$, $\lambda_{\mathrm{deg}}=0.1$ \\
\bottomrule
\end{tabular}
\end{table}

\begin{table}[!htbp]
\centering
\footnotesize
\setlength{\tabcolsep}{4pt}
\renewcommand{\arraystretch}{1.2}

\newcommand{\subcomp}[1]{%
  \hspace*{0.8em}\(\triangleright\)\hspace{0.35em}#1%
}

\caption{Configuration of the~\name\ rectifier, where $S$ is the number
of spectral bands, $S'=4\lceil S/4\rceil$ the zero-padded band count,
and $r$ the MCR bottleneck rank.}
\label{tab:supp_arch}
\begin{tabular}{@{}
    >{\raggedright\arraybackslash}p{0.32\columnwidth}
    >{\raggedright\arraybackslash}p{0.56\columnwidth}
@{}}
\toprule
Component & Configuration \\
\midrule

Spectral groups
& $4$ contiguous groups of $\lceil S/4\rceil$ bands
(bands zero-padded to a multiple of four; padding removed after
back-projection) \\

\addlinespace[2pt]
H-S$^{3}$A blocks
& $4$, applied in sequence with independent parameters and a residual
connection around each:
$F^{(l)}=F^{(l-1)}+\mathrm{Block}_{l}(F^{(l-1)})$, where $F^{(0)}$
is the zero-padded input and $F^{(4)}=F_{s}$ enters MCR \\

\subcomp{per-group stems}
& identity / $3\times3$ / $5\times5$ depthwise /
$3\times3$ dilated (one per group); $1\times1$ mixing after the shuffle \\

\subcomp{TSA views}
& height, width, spectral ($h,w,s$) \\

\addlinespace[2pt]
MCR blocks (refinement stages $N$)
& $1$; ablations that drop MCR remove the module entirely
(projection and back-projection included), so the rectified output is
$\Pi_{S}(F_{s})$, giving the H-S$^{3}$A-only variant of the component
ablation \\

\subcomp{projection $P(\cdot)$}
& $1\times1$ conv, $S'\rightarrow r$, GeLU \\

\subcomp{refine $f_m(\cdot)$}
& $1\times1$ conv at rank $r$, GeLU \\

\subcomp{back-projection $g_m(\cdot)$}
& residual $3\times3$ conv at rank $r$, then $1\times1$,
$r\rightarrow S'$ \\

\addlinespace[2pt]
Bottleneck rank $r$
& $8$ \\

Added parameters
& $+0.048$\,M (independent of the backbone) \\
\bottomrule
\end{tabular}
\end{table}

\section{Test-Time Degradation Protocol}
\label{sec:supp_deg}
In the robustness study, every model is trained once under the standard bicubic degradation and then evaluated without retraining, changing only the test-time degradation. Table~\ref{tab:supp_deg} lists the nine settings, grouped into kernel, blur, noise, and combined mismatch. Blur is applied in the HR space before downsampling; noise is added independently in the LR space after downsampling, using a deterministic per-image seed so that the perturbation is reproducible.

\begin{table}[!htbp]
\centering
\footnotesize
\setlength{\tabcolsep}{8pt}
\renewcommand{\arraystretch}{1.25}

\caption{Test-time degradation settings. Each is applied only at test time to a model trained under standard bicubic downsampling.}
\label{tab:supp_deg}
\begin{tabular}{cll}
\toprule
Tag & Type & Setting \\
\midrule
K1 & Kernel & area-based downsampling \\
K2 & Kernel & alternative bicubic implementation \\
\midrule
B1 & Blur & Gaussian $\sigma=0.6$, kernel $5\times5$ \\
B2 & Blur & Gaussian $\sigma=1.0$, kernel $7\times7$ \\
\midrule
N1 & Noise & Gaussian $\sigma=1/255$ \\
N2 & Noise & Gaussian $\sigma=2/255$ \\
N3 & Noise & Gaussian $\sigma=5/255$ \\
\midrule
J1 & Blur\,+\,Noise & $\sigma=0.6$, $n=2$ \\
J2 & Blur\,+\,Noise & $\sigma=1.0$, $n=5$ \\
\bottomrule
\end{tabular}
\end{table}

\section{Full Robustness Results}
\label{sec:supp_robust}
This section reports the complete robustness study summarized in the main paper, under the nine test-time degradation settings and the training protocol defined in Section~\ref{sec:supp_deg}. Table~\ref{tab:deg_results_arad} lists the full results on ARAD-1K for SwinIR and SwinIR\,+\,\name at the $\times2$, $\times4$, and $\times8$ scales.

\paragraph{Observation}
Across all nine mismatch settings and scales,~\name reduces mSAM in every setting (Table~\ref{tab:deg_results_arad}), while preserving (and frequently improving) spatial metrics.
 The rectified model stays ahead of the Base in every setting, while the margin varies across conditions.

\begin{table}[!htbp]
\centering
\footnotesize
\setlength{\tabcolsep}{3.4pt}
\renewcommand{\arraystretch}{1.1}
\caption{Robustness to mild test-time degradation mismatch on ARAD-1K (SwinIR vs.\ SwinIR\,+\,\name).
We consider nine test-time settings grouped as kernel mismatch (K1: area downsampling; K2: an alternative bicubic implementation),
blur mismatch (B1/B2: HR-space Gaussian blur with $\sigma\in\{0.6,1.0\}$ using $5\times5$ and $7\times7$ kernels before bicubic downsampling),
noise mismatch (N1/N2/N3: i.i.d.\ Gaussian noise added in LR after downsampling with $\sigma=n/255$ for $n\in\{1,2,5\}$),
and combined blur+noise mismatch (J1/J2: $(\sigma,n)\in\{(0.6,2),(1.0,5)\}$).
Gain denotes the change over the baseline (+Ours $-$ Base); since lower mSAM is better, a negative mSAM Gain indicates improvement.}
\label{tab:deg_results_arad}
\begin{adjustbox}{max width=\linewidth,center}
\begin{tabular}{@{}ccc cc cc ccc cc@{}}
\toprule
\multirow{2}{*}{Scale} & \multirow{2}{*}{Metric} & \multirow{2}{*}{Var.}
& \multicolumn{2}{c}{Kernel} & \multicolumn{2}{c}{Blur}
& \multicolumn{3}{c}{Noise} & \multicolumn{2}{c}{Blur$+$Noise} \\
\cmidrule(lr){4-5}\cmidrule(lr){6-7}\cmidrule(lr){8-10}\cmidrule(lr){11-12}
& & & K1 & K2 & B1 & B2 & N1 & N2 & N3 & J1 & J2 \\
\midrule

\multirow{9}{*}{$\times2$}
& \multirow{3}{*}{mPSNR$\up$} & Base  & 46.9930 & 45.2782 & 42.6891 & 39.8390 & 44.1997 & 42.8779 & 39.0326 & 41.2689 & 36.9781 \\
&                       & \ourscell{+Ours} & \ourscell{47.3542} & \ourscell{48.6912} & \ourscell{46.3376} & \ourscell{41.2105} & \ourscell{46.1393} & \ourscell{42.9718} & \ourscell{39.2598} & \ourscell{41.9803} & \ourscell{37.0884} \\
&                       & Gain  & \gainpos{+0.3612} & \gainpos{+3.4130} & \gainpos{+3.6485} & \gainpos{+1.3715} & \gainpos{+1.9396} & \gainpos{+0.0939} & \gainpos{+0.2272} & \gainpos{+0.7114} & \gainpos{+0.1103} \\
\cmidrule(lr){2-12}
& \multirow{3}{*}{mSSIM$\up$} & Base  & 0.9945 & 0.9917 & 0.9859 & 0.9742 & 0.9879 & 0.9806 & 0.9361 & 0.9754 & 0.9200 \\
&                        & \ourscell{+Ours} & \ourscell{0.9949} & \ourscell{0.9968} & \ourscell{0.9934} & \ourscell{0.9810} & \ourscell{0.9892} & \ourscell{0.9807} & \ourscell{0.9408} & \ourscell{0.9766} & \ourscell{0.9270} \\
&                        & Gain & \gainpos{+0.0004} & \gainpos{+0.0051} & \gainpos{+0.0075} & \gainpos{+0.0068} & \gainpos{+0.0013} & \gainpos{+0.0001} & \gainpos{+0.0047} & \gainpos{+0.0012} & \gainpos{+0.0070} \\
\cmidrule(lr){2-12}
& \multirow{3}{*}{mSAM$\down$} & Base  & 0.6412 & 0.7029 & 0.8262 & 1.0349 & 2.2169 & 3.8313 & 7.6378 & 3.8589 & 7.6421 \\
&                       & \ourscell{+Ours} & \ourscell{0.5743} & \ourscell{0.5793} & \ourscell{0.6685} & \ourscell{0.9354} & \ourscell{2.0748} & \ourscell{3.7932} & \ourscell{6.3217} & \ourscell{3.7346} & \ourscell{7.2463} \\
&                       & Gain  & \gainpos{-0.0669} & \gainpos{-0.1236} & \gainpos{-0.1577} & \gainpos{-0.0995} & \gainpos{-0.1421} & \gainpos{-0.0381} & \gainpos{-1.3161} & \gainpos{-0.1243} & \gainpos{-0.3958} \\
\midrule

\multirow{9}{*}{$\times4$}
& \multirow{3}{*}{mPSNR$\up$} & Base  & 38.0659 & 37.3518 & 37.0133 & 36.4237 & 36.9821 & 36.6436 & 35.2832 & 36.3791 & 34.7529 \\
&                       & \ourscell{+Ours} & \ourscell{38.7683} & \ourscell{40.8013} & \ourscell{40.4649} & \ourscell{39.2507} & \ourscell{39.2291} & \ourscell{38.0369} & \ourscell{35.9106} & \ourscell{37.9000} & \ourscell{35.3443} \\
&                       & Gain  & \gainpos{+0.7024} & \gainpos{+3.4495} & \gainpos{+3.4516} & \gainpos{+2.8270} & \gainpos{+2.2470} & \gainpos{+1.3933} & \gainpos{+0.6274} & \gainpos{+1.5209} & \gainpos{+0.5914} \\
\cmidrule(lr){2-12}
& \multirow{3}{*}{mSSIM$\up$} & Base  & 0.9596 & 0.9531 & 0.9498 & 0.9437 & 0.9464 & 0.9395 & 0.9024 & 0.9367 & 0.8948 \\
&                        & \ourscell{+Ours} & \ourscell{0.9670} & \ourscell{0.9748} & \ourscell{0.9731} & \ourscell{0.9672} & \ourscell{0.9618} & \ourscell{0.9464} & \ourscell{0.9086} & \ourscell{0.9452} & \ourscell{0.8970} \\
&                        & Gain  & \gainpos{+0.0074} & \gainpos{+0.0217} & \gainpos{+0.0233} & \gainpos{+0.0235} & \gainpos{+0.0154} & \gainpos{+0.0069} & \gainpos{+0.0062} & \gainpos{+0.0085} & \gainpos{+0.0022} \\
\cmidrule(lr){2-12}
& \multirow{3}{*}{mSAM$\down$} & Base  & 1.3722 & 1.4583 & 1.4760 & 1.5187 & 2.5611 & 3.8740 & 7.0666 & 3.8760 & 7.0590 \\
&                       & \ourscell{+Ours} & \ourscell{1.3204} & \ourscell{1.2944} & \ourscell{1.2709} & \ourscell{1.2949} & \ourscell{2.2280} & \ourscell{3.2619} & \ourscell{6.8138} & \ourscell{3.2247} & \ourscell{6.8102} \\
&                       & Gain  & \gainpos{-0.0518} & \gainpos{-0.1639} & \gainpos{-0.2051} & \gainpos{-0.2238} & \gainpos{-0.3331} & \gainpos{-0.6121} & \gainpos{-0.2528} & \gainpos{-0.6513} & \gainpos{-0.2488} \\
\midrule

\multirow{9}{*}{$\times8$}
& \multirow{3}{*}{mPSNR$\up$} & Base  & 32.6296 & 32.3557 & 32.3143 & 32.2368 & 32.0617 & 31.8439 & 31.1646 & 31.7823 & 31.0966 \\
&                       & \ourscell{+Ours} & \ourscell{33.3445} & \ourscell{34.8300} & \ourscell{34.8294} & \ourscell{34.7736} & \ourscell{34.1853} & \ourscell{33.6229} & \ourscell{31.9019} & \ourscell{33.6025} & \ourscell{31.9108} \\
&                       & Gain  & \gainpos{+0.7149} & \gainpos{+2.4743} & \gainpos{+2.5151} & \gainpos{+2.5368} & \gainpos{+2.1236} & \gainpos{+1.7790} & \gainpos{+0.7373} & \gainpos{+1.8202} & \gainpos{+0.8142} \\
\cmidrule(lr){2-12}
& \multirow{3}{*}{mSSIM$\up$} & Base  & 0.8769 & 0.8698 & 0.8689 & 0.8672 & 0.8608 & 0.8526 & 0.8230 & 0.8518 & 0.8211 \\
&                        & \ourscell{+Ours} & \ourscell{0.8968} & \ourscell{0.9102} & \ourscell{0.9098} & \ourscell{0.9087} & \ourscell{0.8953} & \ourscell{0.8796} & \ourscell{0.8289} & \ourscell{0.8790} & \ourscell{0.8265} \\
&                        & Gain  & \gainpos{+0.0199} & \gainpos{+0.0404} & \gainpos{+0.0409} & \gainpos{+0.0415} & \gainpos{+0.0345} & \gainpos{+0.0270} & \gainpos{+0.0059} & \gainpos{+0.0272} & \gainpos{+0.0054} \\
\cmidrule(lr){2-12}
& \multirow{3}{*}{mSAM$\down$} & Base  & 2.3049 & 2.4180 & 2.4190 & 2.4220 & 3.0980 & 4.0295 & 6.5479 & 4.0262 & 6.5525 \\
&                       & \ourscell{+Ours} & \ourscell{2.2712} & \ourscell{2.1173} & \ourscell{2.1052} & \ourscell{2.0901} & \ourscell{2.8654} & \ourscell{3.5249} & \ourscell{6.3644} & \ourscell{3.5022} & \ourscell{5.9158} \\
&                       & Gain  & \gainpos{-0.0337} & \gainpos{-0.3007} & \gainpos{-0.3138} & \gainpos{-0.3319} & \gainpos{-0.2326} & \gainpos{-0.5046} & \gainpos{-0.1835} & \gainpos{-0.5240} & \gainpos{-0.6367} \\
\bottomrule
\end{tabular}
\end{adjustbox}
\end{table}

\section{Full Cross-Domain Results}
\label{sec:supp_cross}
Table~\ref{tab:supp_cross_domain} reports the complete cross-domain results at the $\times2$ and $\times4$ scales (stacked as two blocks), including the mSSIM columns omitted from the main paper for brevity. DDS2M is reported in the in-domain setting only, as its diffusion inference is far more costly than the other backbones. The four models in the table are trained on the 900 training images of ARAD-1K and directly evaluated on all CAVE scenes and all ICVL HSIs without fine-tuning. As in the main paper, attaching~\name yields mSAM reductions in every reported configuration on both target domains while preserving (and frequently improving) the spatial metrics.

\begin{table}[!htbp]
\centering
\footnotesize
\setlength{\tabcolsep}{9.5pt}
\renewcommand{\arraystretch}{1.1}
\caption{\looseness=-1 Full cross-domain results at the $\times2$ and $\times4$ scales (trained on ARAD-1K, tested on CAVE and ICVL without fine-tuning), grouped by backbone and scale.
Higher mPSNR and mSSIM and lower mSAM are better; the Gain row is reported as (+Ours $-$ Base), so a negative mSAM gain denotes improved spectral fidelity.
Base denotes each backbone trained on ARAD-1K and directly evaluated on CAVE and ICVL, and +Ours denotes the same backbone augmented with~\name under the identical protocol.}
\label{tab:supp_cross_domain}
\begin{adjustbox}{max width=\linewidth,center}
\begin{tabular}{@{}lcc ccc ccc@{}}
\toprule
\multirow{2}{*}{Method} & \multirow{2}{*}{Scale} & \multirow{2}{*}{Variant}
& \multicolumn{3}{c}{CAVE} & \multicolumn{3}{c}{ICVL} \\
\cmidrule(lr){4-6}\cmidrule(lr){7-9}
& & & mPSNR\Up & mSSIM\Up & mSAM\Down & mPSNR\Up & mSSIM\Up & mSAM\Down \\
\midrule
\multirow{6}{*}{RCAN}
& \multirow{3}{*}{$\times2$} & Base & 42.7269 & 0.9866 & 5.0004 & 49.6448 & 0.9976 & 0.7389 \\
& & \ourscell{+Ours} & \ourscell{44.2866} & \ourscell{0.9886} & \ourscell{4.0978} & \ourscell{51.8506} & \ourscell{0.9979} & \ourscell{0.4682} \\
& & Gain & \gainpos{+1.5597} & \gainpos{+0.0020} & \gainpos{-0.9026} & \gainpos{+2.2058} & \gainpos{+0.0003} & \gainpos{-0.2707} \\
\cmidrule(lr){2-9}
& \multirow{3}{*}{$\times4$} & Base & 35.6214 & 0.9537 & 6.2153 & 39.9389 & 0.9707 & 1.0969 \\
& & \ourscell{+Ours} & \ourscell{36.3326} & \ourscell{0.9576} & \ourscell{5.3594} & \ourscell{40.9148} & \ourscell{0.9723} & \ourscell{0.7897} \\
& & Gain & \gainpos{+0.7112} & \gainpos{+0.0039} & \gainpos{-0.8559} & \gainpos{+0.9759} & \gainpos{+0.0016} & \gainpos{-0.3072} \\
\midrule
\multirow{6}{*}{SSPSR}
& \multirow{3}{*}{$\times2$} & Base & 41.4071 & 0.9816 & 4.0419 & 48.4758 & 0.9966 & 0.8140 \\
& & \ourscell{+Ours} & \ourscell{43.0076} & \ourscell{0.9845} & \ourscell{2.9243} & \ourscell{50.7217} & \ourscell{0.9972} & \ourscell{0.5264} \\
& & Gain & \gainpos{+1.6005} & \gainpos{+0.0029} & \gainpos{-1.1176} & \gainpos{+2.2459} & \gainpos{+0.0006} & \gainpos{-0.2876} \\
\cmidrule(lr){2-9}
& \multirow{3}{*}{$\times4$} & Base & 36.7963 & 0.9521 & 4.9284 & 43.9812 & 0.9851 & 1.0271 \\
& & \ourscell{+Ours} & \ourscell{37.5721} & \ourscell{0.9557} & \ourscell{3.9189} & \ourscell{45.1945} & \ourscell{0.9856} & \ourscell{0.6698} \\
& & Gain & \gainpos{+0.7758} & \gainpos{+0.0036} & \gainpos{-1.0095} & \gainpos{+1.2133} & \gainpos{+0.0005} & \gainpos{-0.3573} \\
\midrule
\multirow{6}{*}{SwinIR}
& \multirow{3}{*}{$\times2$} & Base & 40.7157 & 0.9839 & 4.4366 & 47.9367 & 0.9964 & 0.8260 \\
& & \ourscell{+Ours} & \ourscell{42.3415} & \ourscell{0.9877} & \ourscell{2.6231} & \ourscell{50.0704} & \ourscell{0.9971} & \ourscell{0.4709} \\
& & Gain & \gainpos{+1.6258} & \gainpos{+0.0038} & \gainpos{-1.8135} & \gainpos{+2.1337} & \gainpos{+0.0007} & \gainpos{-0.3551} \\
\cmidrule(lr){2-9}
& \multirow{3}{*}{$\times4$} & Base & 34.7725 & 0.9313 & 4.8551 & 41.0743 & 0.9708 & 1.0552 \\
& & \ourscell{+Ours} & \ourscell{35.7256} & \ourscell{0.9577} & \ourscell{3.8360} & \ourscell{42.6530} & \ourscell{0.9799} & \ourscell{0.6682} \\
& & Gain & \gainpos{+0.9531} & \gainpos{+0.0264} & \gainpos{-1.0191} & \gainpos{+1.5787} & \gainpos{+0.0091} & \gainpos{-0.3870} \\
\midrule
\multirow{6}{*}{ESSAformer}
& \multirow{3}{*}{$\times2$} & Base & 42.1763 & 0.9835 & 3.3575 & 49.7895 & 0.9973 & 0.6404 \\
& & \ourscell{+Ours} & \ourscell{43.4978} & \ourscell{0.9855} & \ourscell{2.5611} & \ourscell{51.3981} & \ourscell{0.9977} & \ourscell{0.4625} \\
& & Gain & \gainpos{+1.3215} & \gainpos{+0.0020} & \gainpos{-0.7964} & \gainpos{+1.6086} & \gainpos{+0.0004} & \gainpos{-0.1779} \\
\cmidrule(lr){2-9}
& \multirow{3}{*}{$\times4$} & Base & 37.0305 & 0.9556 & 4.8258 & 44.3869 & 0.9856 & 0.9349 \\
& & \ourscell{+Ours} & \ourscell{37.6350} & \ourscell{0.9579} & \ourscell{3.5556} & \ourscell{45.7012} & \ourscell{0.9866} & \ourscell{0.6079} \\
& & Gain & \gainpos{+0.6045} & \gainpos{+0.0023} & \gainpos{-1.2702} & \gainpos{+1.3143} & \gainpos{+0.0010} & \gainpos{-0.3270} \\
\bottomrule
\end{tabular}
\end{adjustbox}
\end{table}

\section{Rectifier Controls and Seed Stability: Implementation and Full Results}
\label{sec:supp_simple}
The main paper reports the $\times4$ subset of the comparison against simple spectral rectifiers. Here we provide the full three-scale comparison and the implementation details of the fixed post-hoc rectifiers. All three baselines operate on the output of a fixed SwinIR model and do not modify the SR backbone weights, whereas~\name is trained end-to-end as a model-agnostic rectification module.

\paragraph{Fixed rectifiers}
\textbf{Savitzky--Golay smoothing (SG)} applies a 1D Savitzky--Golay filter along the spectral dimension for each pixel independently, with window length $w=7$ and polynomial order $p=3$;
the spectrum is symmetrically reflection-padded before filtering and cropped back to avoid boundary bias.
\textbf{PCA projection} fits a dataset-level PCA subspace from $20{,}000$ pixel spectra randomly sampled from the ARAD-1K training split and keeps it fixed at test time;
with rank $k=8$, each estimated spectrum is projected onto the top-$k$ principal components and reconstructed back to the original spectral space.
\textbf{Iterative back-projection (IBP)} refines the SwinIR output by enforcing consistency with the observed LR image under the assumed degradation, performing $T=10$ steps with step size $\eta=1.0$,
\begin{equation}
\textit{I}_{\mathrm{SR}}^{(t+1)}
=
\textit{I}_{\mathrm{SR}}^{(t)}
+
\eta \cdot \mathcal{U}
\!\left(
\textit{I}_{\mathrm{LR}}
-
\mathcal{D}\!\left(\textit{I}_{\mathrm{SR}}^{(t)}\right)
\right),
\end{equation}
where $\mathcal{D}$ is the same bicubic downsampling operator used in training and $\mathcal{U}$ is its bicubic upsampling counterpart, applied independently per band to keep the procedure lightweight.

\paragraph{Learned controls}
Two learned baselines isolate the source of the gain, both trained jointly with SwinIR under the identical loss, cosine schedule, $1000$ epochs, and seed as~\name at $\times4$ on ARAD-1K.
\textbf{$\mathcal{L}_{\mathrm{deg}}$ only} adds the degradation-consistency constraint to the Base with no rectifier module (no extra parameters).
\textbf{Generic head} is a same-budget rectifier that replaces H-S$^{3}$A and MCR by a plain residual block of three $3\times3$ convolutions ($31\!\to\!48\!\to\!48\!\to\!31$ channels with GeLU activations) and a global residual connection, adding $+0.048$M parameters to match~\name; unlike MCR it has no low-dimensional spectral bottleneck. These two learned controls are evaluated at $\times4$ only.

\paragraph{Seed stability}
To confirm the attribution is robust, we retrain the five SwinIR/ARAD-1K variants with three seeds ($\{42,123,2024\}$) and report the mean$\,\pm\,$std in Table~\ref{tab:seed}. The across-seed standard deviation is small ($\le0.012$ mSAM), and the gap between the cheap controls ($1.362$ and $1.371$ mSAM) and MCR\,/\,\name ($1.319$ and $1.280$) ranges from $0.043$ to $0.091$, at least $3.5\times$ that deviation, so the ranking, and the conclusion that the gain comes from the architecture, is stable across seeds.

\begin{table}[!htbp]
\centering
\footnotesize
\setlength{\tabcolsep}{2.7pt}
\renewcommand{\arraystretch}{1.1}
\caption{Spectral-rectifier comparison on ARAD-1K with SwinIR. SG, PCA, and IBP are fixed post-hoc methods; MCR and~\name are jointly trained. Best in \textbf{bold}, second best \underline{underlined}.}
\label{tab:supp_simple_rectifiers}
\begin{tabular}{@{}lccccc@{}}
\toprule
Method & Scale & mPSNR$\uparrow$ & mSAM$\downarrow$ & mSSIM$\uparrow$ & CC$\uparrow$ \\
\midrule
\multirow{3}{*}{Base (SwinIR)}
& $\times2$ & 50.4662 & 0.7727 & 0.9972 & 0.9990 \\
& $\times4$ & 39.5717 & 1.3950 & 0.9709 & 0.9902 \\
& $\times8$ & 33.4185 & 2.4682 & 0.8869 & 0.9551 \\
\midrule
\multirow{3}{*}{Base + SG}
& $\times2$ & 46.1893 & 1.9059 & 0.9959 & 0.9975 \\
& $\times4$ & 38.7834 & 2.4512 & 0.9698 & 0.9889 \\
& $\times8$ & 33.2419 & 3.2638 & 0.8859 & 0.9541 \\
\midrule
\multirow{3}{*}{Base + PCA}
& $\times2$ & 50.7837 & 0.6188 & 0.9972 & 0.9990 \\
& $\times4$ & 39.5729 & 1.4028 & 0.9709 & 0.9902 \\
& $\times8$ & 33.4198 & 2.4601 & 0.8869 & 0.9552 \\
\midrule
\multirow{3}{*}{Base + IBP}
& $\times2$ & 51.0962 & 0.5211 & \underline{0.9973} & \underline{0.9992} \\
& $\times4$ & 39.5782 & 1.3693 & 0.9709 & 0.9902 \\
& $\times8$ & 33.4041 & 2.4365 & 0.8865 & 0.9548 \\
\midrule
\multirow{3}{*}{Base + MCR}
& $\times2$ & \underline{51.2531} & \underline{0.5182} & \underline{0.9973} & \underline{0.9992} \\
& $\times4$ & \underline{40.2550} & \underline{1.3173} & \underline{0.9717} & \underline{0.9912} \\
& $\times8$ & \underline{33.9241} & \underline{2.2159} & \underline{0.8947} & \underline{0.9675} \\
\midrule
\multirow{3}{*}{Base + Ours}
& \cellcolor{oursrow}$\times2$ & \cellcolor{oursrow}\textbf{51.6551} & \cellcolor{oursrow}\textbf{0.4842} & \cellcolor{oursrow}\textbf{0.9975} & \cellcolor{oursrow}\textbf{0.9993} \\
& \cellcolor{oursrow}$\times4$ & \cellcolor{oursrow}\textbf{40.9720} & \cellcolor{oursrow}\textbf{1.2819} & \cellcolor{oursrow}\textbf{0.9752} & \cellcolor{oursrow}\textbf{0.9924} \\
& \cellcolor{oursrow}$\times8$ & \cellcolor{oursrow}\textbf{34.8318} & \cellcolor{oursrow}\textbf{2.1164} & \cellcolor{oursrow}\textbf{0.9103} & \cellcolor{oursrow}\textbf{0.9720} \\
\bottomrule
\end{tabular}
\end{table}

\begin{table}[!htbp]
\centering
\footnotesize
\setlength{\tabcolsep}{2pt}
\renewcommand{\arraystretch}{1.1}
\caption{Seed stability on ARAD-1K ($\times4$, SwinIR): mean$\,\pm\,$std over three seeds $\{42,123,2024\}$. The architecture gap (cheap controls versus MCR/\name) is $0.043$ to $0.091$ mSAM, at least \mbox{$3.5\times$ the across-seed std.}}
\label{tab:seed}
\begin{tabular}{@{}lccc@{}}
\toprule
Variant (SwinIR, $\times4$) & mPSNR$\uparrow$ & mSAM$\downarrow$ & mSSIM$\uparrow$ \\
\midrule
Base            & $39.58_{\pm.03}$ & $1.397_{\pm.011}$ & $0.9710_{\pm.0005}$ \\
\;+ $\mathcal{L}_{\mathrm{deg}}$ only & $39.70_{\pm.03}$ & $1.371_{\pm.010}$ & $0.9710_{\pm.0007}$ \\
\;+ Generic head & $39.99_{\pm.04}$ & $1.362_{\pm.012}$ & $0.9713_{\pm.0006}$ \\
\;+ MCR only     & $40.25_{\pm.03}$ & $1.319_{\pm.010}$ & $0.9716_{\pm.0006}$ \\
\rowcolor{oursrow}\;+~\name        & $\mathbf{40.96}_{\pm.03}$ & $\mathbf{1.280}_{\pm.009}$ & $\mathbf{0.9753}_{\pm.0006}$ \\
\bottomrule
\end{tabular}
\end{table}

\begin{table}[!htbp]
\centering
\footnotesize
\setlength{\tabcolsep}{5.0pt}
\renewcommand{\arraystretch}{1.1}
\caption{Complexity of~\name using SwinIR on ARAD-1K at $\times4$. Parameters and FLOPs are computed for the $32^2$ LR training patch ($128^2$ HR); the rectifier adds $0.048$M parameters \mbox{and $1.482$G FLOPs.}}
\label{tab:complexity_loss}
\begin{tabular}{@{}cc ccc@{}}
\toprule
Scale & Metric & Base & \cellcolor{oursrow}+Ours & $\Delta$ \\
\midrule
\multirow{2}{*}{$\times4$}
 & Params (M) & 8.228  & \ourscell{8.276}  & \gainpos{+0.048} \\
 & FLOPs (G)  & 19.302 & \ourscell{20.785} & \gainpos{+1.482} \\
\bottomrule
\end{tabular}
\end{table}

\section{Structural Hyperparameters of the Rectifier}
\label{sec:supp_struct}
\looseness=-1 We vary the three structural hyperparameters of~\name one at a time on ARAD-1K at $\times4$ with SwinIR, keeping every other setting at its default ($r=8$, $4$ H-S$^{3}$A blocks, $N=1$ MCR block, $4$ spectral groups, $\lambda_{\mathrm{deg}}=0.1$). Table~\ref{tab:struct} reports the sweeps. The default row of each sweep is the full model of the main paper, and the $0$-block and w/o-MCR rows are its single-stage ablations; these rows quote the main paper's values. The w/o-MCR row removes the module entirely, projection and back-projection included; it is not obtained by truncating the correction sum at $i=0$, which would retain the term $g_m(m_0)$.

The bottleneck rank behaves as a two-sided constraint. Setting $r=1$ raises mSAM to $1.349$, and widening the projection to the band count ($r=31$, one below its $S'=32$ input) raises it to $1.331$, against $1.2819$ at the default; both margins exceed the across-seed standard deviation of Section~\ref{sec:supp_simple} ($\le0.012$ mSAM) by more than fourfold. The $r=31$ variant has more parameters than the default, so the loss is not a capacity effect. Between $r=4$ and $r=16$ mSAM varies by $0.006$, which is below that standard deviation, so the width of the bottleneck is not critical in this range.

Both depth sweeps flatten at or before the default. A single H-S$^{3}$A block leaves mSAM at $1.322$, close to the $1.3173$ obtained with none; two and four blocks reach $1.298$ and $1.2819$, and eight give $1.288$. For MCR, $N=2$ is nominally the best setting ($1.278$ against $1.2819$ at $N=1$), while $N=4$ and $N=8$ fall back to $1.289$ and $1.294$; every $N\ge1$ lies within $0.013$ of the default, and we use $N=1$, the cheapest of them.

Disabling the adjacent spectra shuffle while keeping the $1\times1$ mixing leaves the parameter count unchanged but raises mSAM from $1.2819$ to $1.3102$, more than twice the across-seed standard deviation, while mPSNR moves only from $40.9720$ to $40.8871$. Compared with the MCR-only variant ($1.3173$), the shuffle thus carries most of the spectral contribution of the H-S$^{3}$A stage, whereas the spatial gain largely persists without it. Because the fixed permutation is immediately followed by a dense $1\times1$ convolution, it does not enlarge the function class; its benefit is an optimization effect of the fixed adjacent-group coupling.

\begin{table}[!htbp]
\centering
\captionsetup{skip=3pt}
\footnotesize
\setlength{\tabcolsep}{1.0pt}
\renewcommand{\arraystretch}{1.1}

\caption{Structural hyperparameter sweeps on ARAD-1K ($\times4$, SwinIR). Each block varies one hyperparameter of the rectifier while every other setting keeps the default of the main paper ($r=8$, four H-S$^{3}$A blocks, $N=1$ MCR block, adjacent spectra shuffle on); $^*$ marks that default, so all $^*$ rows are the same full model. Best in \textbf{bold}, second best \underline{underlined} within each block; the across-seed standard deviation is $\le0.012$ mSAM (Table~\ref{tab:seed}).}
\label{tab:struct}
\begin{minipage}[t]{0.46\textwidth}
\centering
\begin{adjustbox}{max width=\linewidth,center}
\begin{tabular}[t]{llccc}
\toprule
Sweep & Setting & mPSNR$\uparrow$ & mSSIM$\uparrow$ & mSAM$\downarrow$ \\
\midrule
\multirow{6}{*}{MCR rank $r$}
 & $1$ & 40.44 & 0.9726 & 1.349 \\
 & $2$ & 40.67 & 0.9738 & 1.326 \\
 & $4$ & 40.91 & \underline{0.9748} & 1.288 \\
 & \cellcolor{oursrow}$8^{*}$ & \cellcolor{oursrow}\textbf{40.97} & \cellcolor{oursrow}\textbf{0.9752} & \cellcolor{oursrow}\textbf{1.282} \\
 & $16$ & \underline{40.92} & 0.9745 & \underline{1.284} \\
 & $31$ & 40.82 & 0.9741 & 1.331 \\
\midrule
\multirow{5}{*}{H-S$^{3}$A blocks}
 & $0$ (MCR only) & 40.26 & 0.9717 & 1.317 \\
 & $1$ & 40.58 & 0.9727 & 1.322 \\
 & $2$ & 40.84 & 0.9743 & 1.298 \\
 & \cellcolor{oursrow}$4^{*}$ & \cellcolor{oursrow}\textbf{40.97} & \cellcolor{oursrow}\textbf{0.9752} & \cellcolor{oursrow}\textbf{1.282} \\
 & $8$ & \underline{40.93} & \underline{0.9747} & \underline{1.288} \\
\bottomrule
\end{tabular}
\end{adjustbox}
\end{minipage}\hfill
\begin{minipage}[t]{0.46\textwidth}
\centering
\begin{adjustbox}{max width=\linewidth,center}
\begin{tabular}[t]{llccc}
\toprule
Sweep & Setting & mPSNR$\uparrow$ & mSSIM$\uparrow$ & mSAM$\downarrow$ \\
\midrule
\multirow{5}{*}{MCR blocks $N$}
 & w/o (H-S$^{3}$A only) & 40.71 & 0.9735 & 1.348 \\
 & \cellcolor{oursrow}$1^{*}$ & \cellcolor{oursrow}\underline{40.97} & \cellcolor{oursrow}0.9752 & \cellcolor{oursrow}\underline{1.282} \\
 & $2$ & \textbf{40.99} & \textbf{0.9754} & \textbf{1.278} \\
 & $4$ & 40.97 & \underline{0.9753} & 1.289 \\
 & $8$ & 40.91 & 0.9746 & 1.294 \\
\midrule
\multirow{2}{*}{Spectra shuffle}
 & \cellcolor{oursrow}with$^{*}$ & \cellcolor{oursrow}\textbf{40.97} & \cellcolor{oursrow}\textbf{0.9752} & \cellcolor{oursrow}\textbf{1.282} \\
 & w/o & 40.89 & 0.9743 & 1.310 \\
\bottomrule
\end{tabular}
\end{adjustbox}
\end{minipage}
\end{table}

\section{Dataset-Level Spectral-Error and Significance Analysis}
\label{sec:supp_significance}
To complement the per-pixel spectral profiles in the main paper, we quantify the spectral correction over the full ARAD-1K test set (50 scenes) at $\times4$ with SwinIR. For each pixel we form the residual spectrum $e=\hat{\textit{I}}_{\mathrm{SR}}-\textit{I}_{\mathrm{GT}}$ and report its mean absolute value (MAE), its mean absolute first spectral difference $|\Delta e|$, and its mean absolute second spectral difference $|\Delta^{2} e|$ (a proxy for the band-to-band roughness of the error). As summarized in Table~\ref{tab:significance},~\name lowers all three statistics: the error magnitude on every scene, and its first- and second-difference variation on 44 and 42 of 50, respectively; per-scene mSAM improves on all 50 scenes. A one-sided Wilcoxon signed-rank test finds every difference significant. This shows at the dataset level that the improvement is systematic rather than driven by a few scenes. The three statistics fall at different rates: $|\Delta^{2} e|$ decreases in absolute terms together with the error magnitude, but relative to the (smaller) residual it increases.

\begin{table}[!htbp]
\centering
\footnotesize
\setlength{\tabcolsep}{4.5pt}
\renewcommand{\arraystretch}{1.1}
\caption{Dataset-level residual spectral-error analysis over 50 ARAD-1K test scenes (SwinIR, $\times4$). ``Improved'' counts scenes on which~\name lowers the statistic; $p$ is from a one-sided Wilcoxon signed-rank test. Per-scene mSAM improves on all 50 scenes ($p<10^{-6}$).}
\label{tab:significance}
\begin{tabular}{@{}lcccc@{}}
\toprule
Residual statistic & Base & \cellcolor{oursrow}+\,\name & Improved & $p$ \\
\midrule
Spectral MAE                 & 0.00783  & \cellcolor{oursrow}\textbf{0.00534}  & 50/50 & $<10^{-6}$ \\
Gradient $|\Delta e|$        & 0.000838 & \cellcolor{oursrow}\textbf{0.000781} & 44/50 & $<10^{-3}$ \\
Curvature $|\Delta^{2} e|$   & 0.000697 & \cellcolor{oursrow}\textbf{0.000618} & 42/50 & $<10^{-3}$ \\
\bottomrule
\end{tabular}
\end{table}

\begin{table}[!htbp]
\centering
\footnotesize
\setlength{\tabcolsep}{4.2pt}
\renewcommand{\arraystretch}{1.1}
\captionsetup{skip=3pt}
\caption{Frozen-backbone results. Each backbone is held at its Base weights and only~\name is trained on top of it, so the backbone is used exactly as it was originally trained. mSAM falls in all sixteen configurations, recovering $76\%$ of the jointly trained gain reported in the main paper on average, with per-configuration recovery ranging from $71\%$ to $87\%$. Across the remaining metrics the single exception is ESSAformer's mSSIM on ARAD-1K at $\times2$, which changes by $-0.0001$.}
\label{tab:frozen_backbone}
\begin{adjustbox}{max width=\linewidth,center}
\begin{tabular}{@{}ll cP cP cP cP@{}}
\toprule
 & & \multicolumn{2}{c}{RCAN} & \multicolumn{2}{c}{SSPSR} & \multicolumn{2}{c}{SwinIR} & \multicolumn{2}{c@{}}{ESSAformer} \\
\cmidrule(lr){3-4}\cmidrule(lr){5-6}\cmidrule(lr){7-8}\cmidrule(l){9-10}
Setting & Metric & Base & Frozen & Base & Frozen & Base & Frozen & Base & Frozen \\
\midrule
\multirow{4}{*}{ARAD-1K $\times2$}
 & mPSNR$\uparrow$   & 48.74 & 49.16 & 46.09 & 46.49 & 50.47 & 50.64 & 49.45 & 49.61 \\
 & mSSIM$\uparrow$   & 0.9936 & 0.9948 & 0.9893 & 0.9894 & 0.9972 & 0.9974 & 0.9937 & 0.9936 \\
 & mSAM$\downarrow$  & 1.2071 & 0.9043 & 0.9906 & 0.7541 & 0.7727 & 0.5642 & 0.6090 & 0.5321 \\
 & CC$\uparrow$      & 0.9978 & 0.9984 & 0.9960 & 0.9979 & 0.9990 & 0.9991 & 0.9986 & 0.9989 \\
\midrule
\multirow{4}{*}{ARAD-1K $\times4$}
 & mPSNR$\uparrow$   & 38.38 & 38.65 & 38.88 & 39.05 & 39.57 & 39.76 & 39.95 & 40.08 \\
 & mSSIM$\uparrow$   & 0.9470 & 0.9536 & 0.9463 & 0.9483 & 0.9709 & 0.9725 & 0.9592 & 0.9595 \\
 & mSAM$\downarrow$  & 1.9009 & 1.4622 & 1.6076 & 1.3965 & 1.3950 & 1.3142 & 1.3670 & 1.2682 \\
 & CC$\uparrow$      & 0.9812 & 0.9895 & 0.9858 & 0.9875 & 0.9902 & 0.9914 & 0.9903 & 0.9906 \\
\midrule
\multirow{4}{*}{CAVE $\times2$}
 & mPSNR$\uparrow$   & 41.09 & 41.72 & 39.40 & 39.71 & 40.31 & 40.68 & 39.47 & 39.86 \\
 & mSSIM$\uparrow$   & 0.9838 & 0.9863 & 0.9711 & 0.9748 & 0.9754 & 0.9786 & 0.9704 & 0.9708 \\
 & mSAM$\downarrow$  & 5.7619 & 4.7216 & 4.2518 & 3.8642 & 0.8214 & 0.6723 & 4.0820 & 3.4925 \\
 & CC$\uparrow$      & 0.9946 & 0.9965 & 0.9907 & 0.9942 & 0.9807 & 0.9882 & 0.9950 & 0.9954 \\
\midrule
\multirow{4}{*}{CAVE $\times4$}
 & mPSNR$\uparrow$   & 35.99 & 36.42 & 35.24 & 35.41 & 37.27 & 37.51 & 34.11 & 34.48 \\
 & mSSIM$\uparrow$   & 0.9536 & 0.9572 & 0.9384 & 0.9401 & 0.9492 & 0.9527 & 0.9267 & 0.9274 \\
 & mSAM$\downarrow$  & 7.6445 & 6.0418 & 4.8735 & 4.4726 & 0.9015 & 0.7582 & 5.6070 & 4.7213 \\
 & CC$\uparrow$      & 0.9825 & 0.9853 & 0.9795 & 0.9808 & 0.9315 & 0.9461 & 0.9846 & 0.9873 \\
\bottomrule
\end{tabular}
\end{adjustbox}
\end{table}

\clearpage
\raggedbottom
\section{Additional Error-Map Galleries}
\label{sec:supp_errmap}
Figures~\ref{fig:errmap_arad}--\ref{fig:errmap_icvl} provide per-scene error-map galleries at $\times4$: on ARAD-1K and CAVE under the in-domain setting, and on ICVL under the cross-domain setting, where models are trained on ARAD-1K and tested on ICVL without fine-tuning. For each scene we show pseudo-RGB renderings and the band-averaged absolute error maps of different SR backbones, with each rectified result (+Ours) placed next to its baseline. On the unseen ICVL data, the boundary-related errors are reduced in the same way as in the in-domain comparisons of the main paper.

\ErrMapGalleryBlock
{Error-map gallery on ARAD-1K at $\times4$ (in-domain).}
{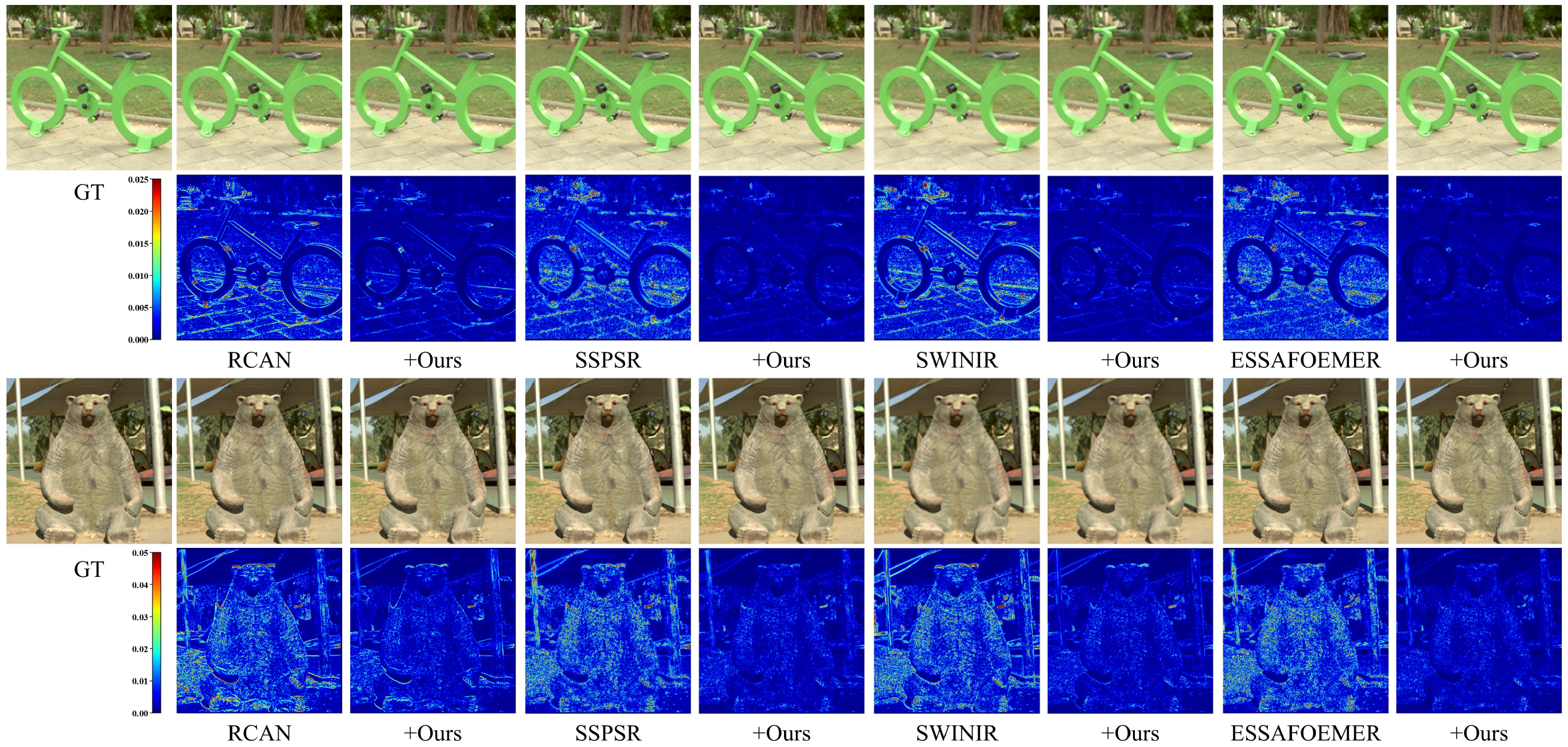}{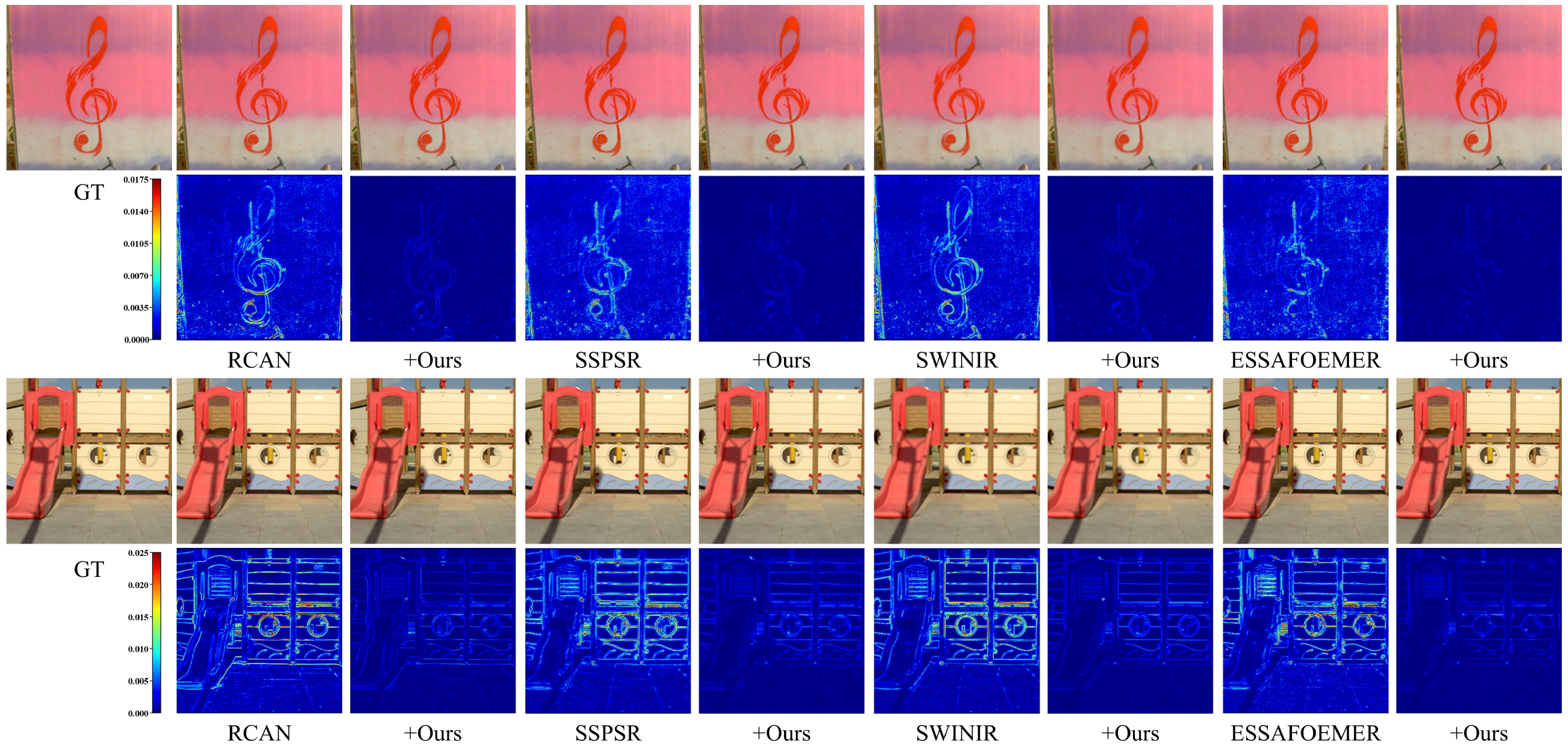}{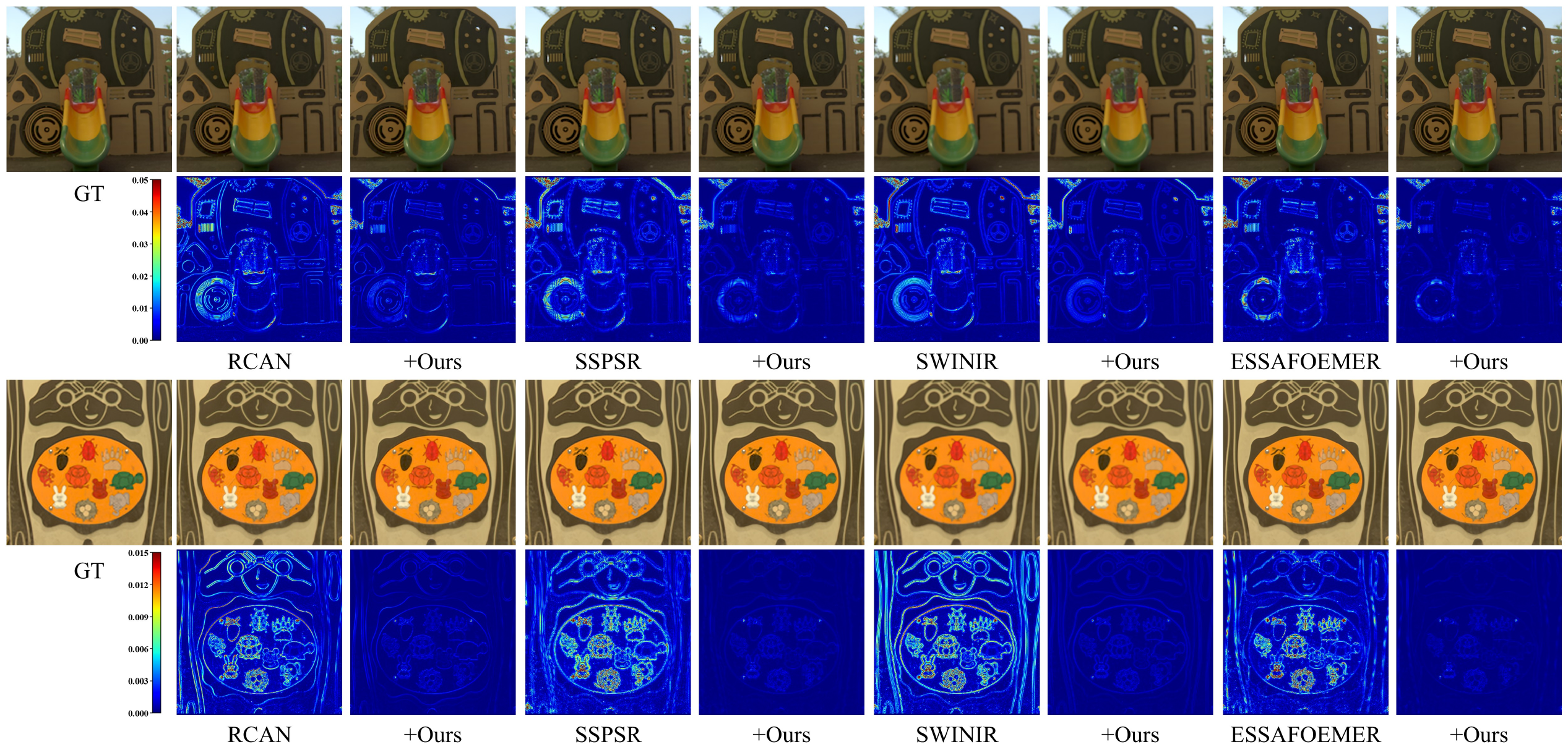}{fig:errmap_arad}{0.255}

\clearpage
\ErrMapGalleryBlock
{Error-map gallery on CAVE at $\times4$ (in-domain). The visualization protocol is \mbox{identical to Figure~\ref{fig:errmap_arad}.}}
{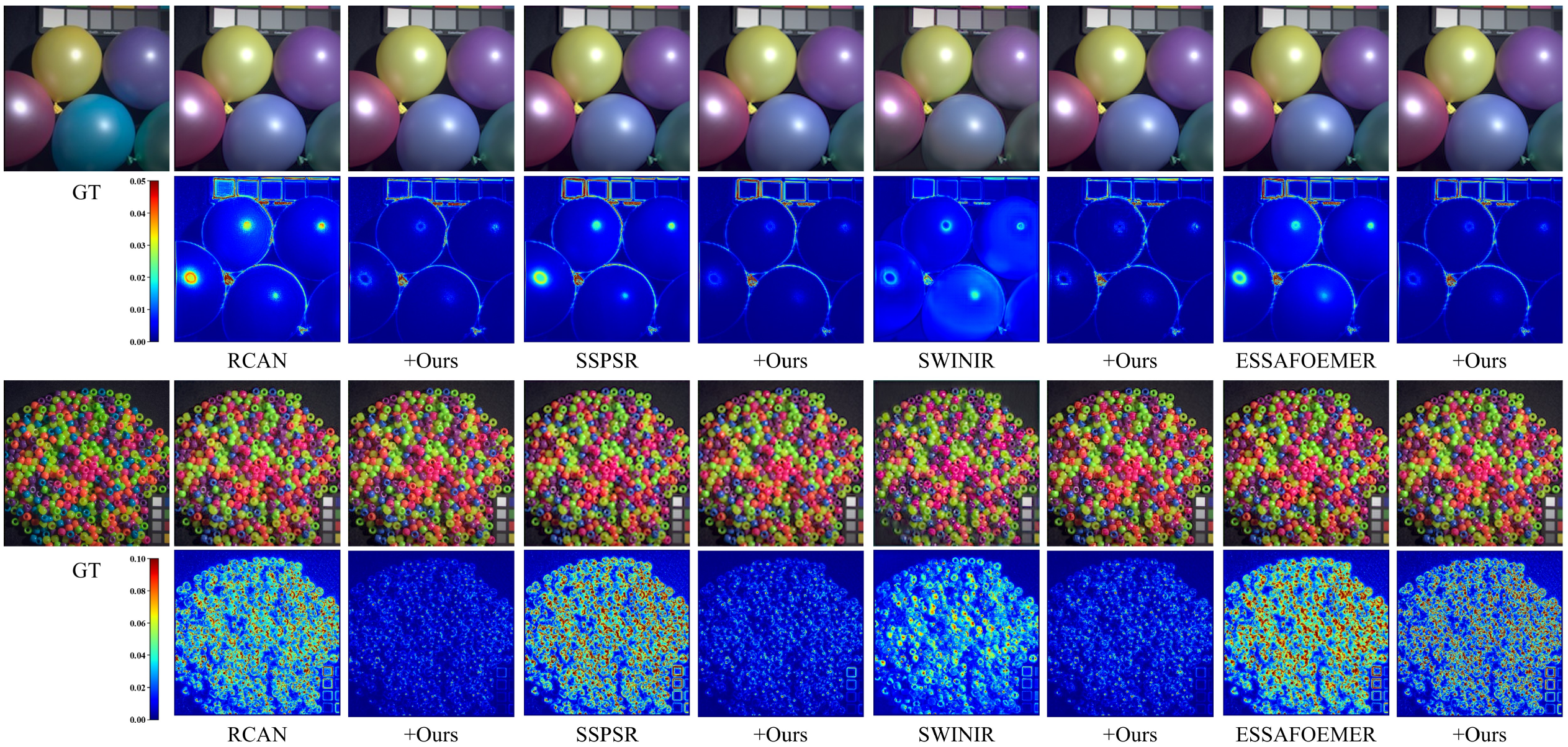}{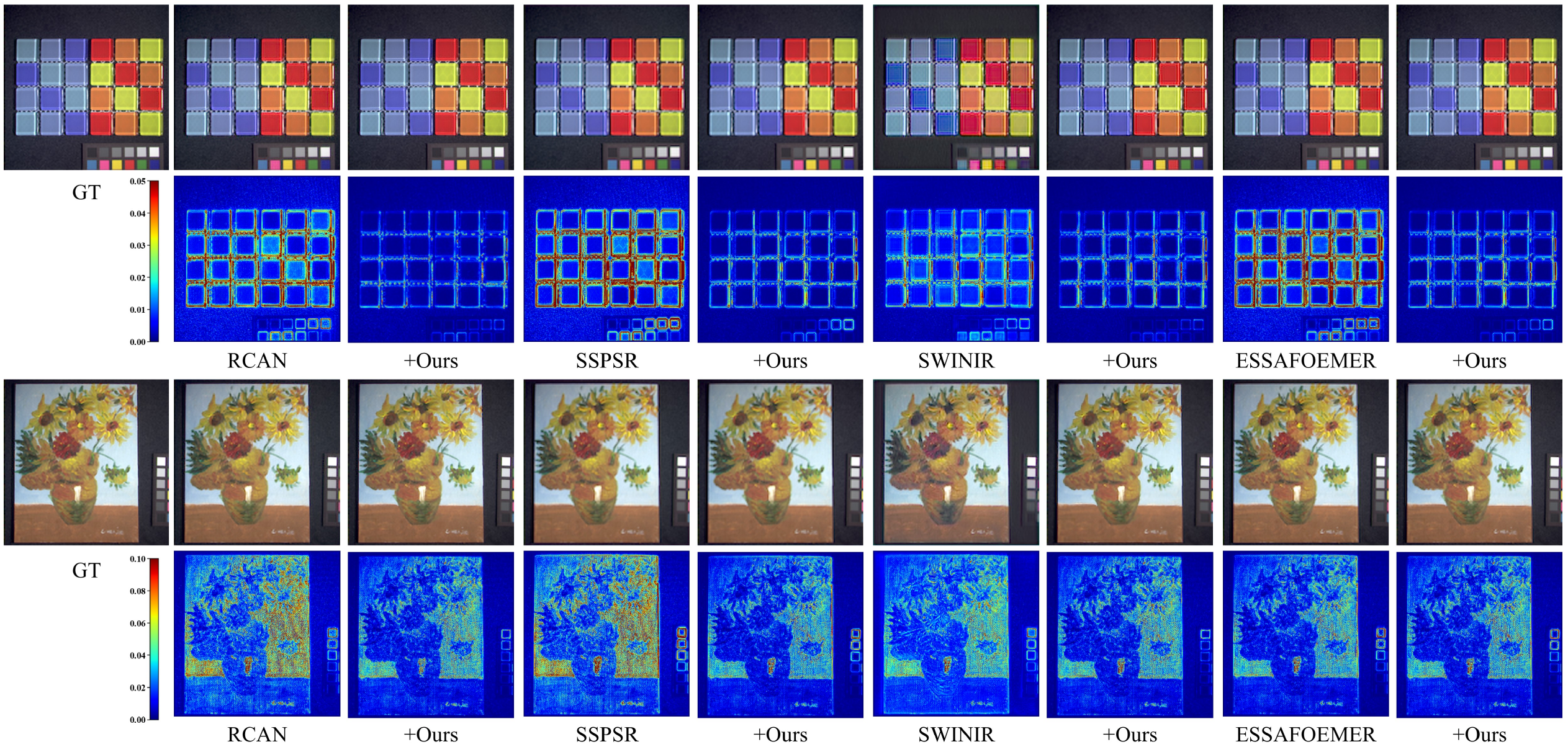}{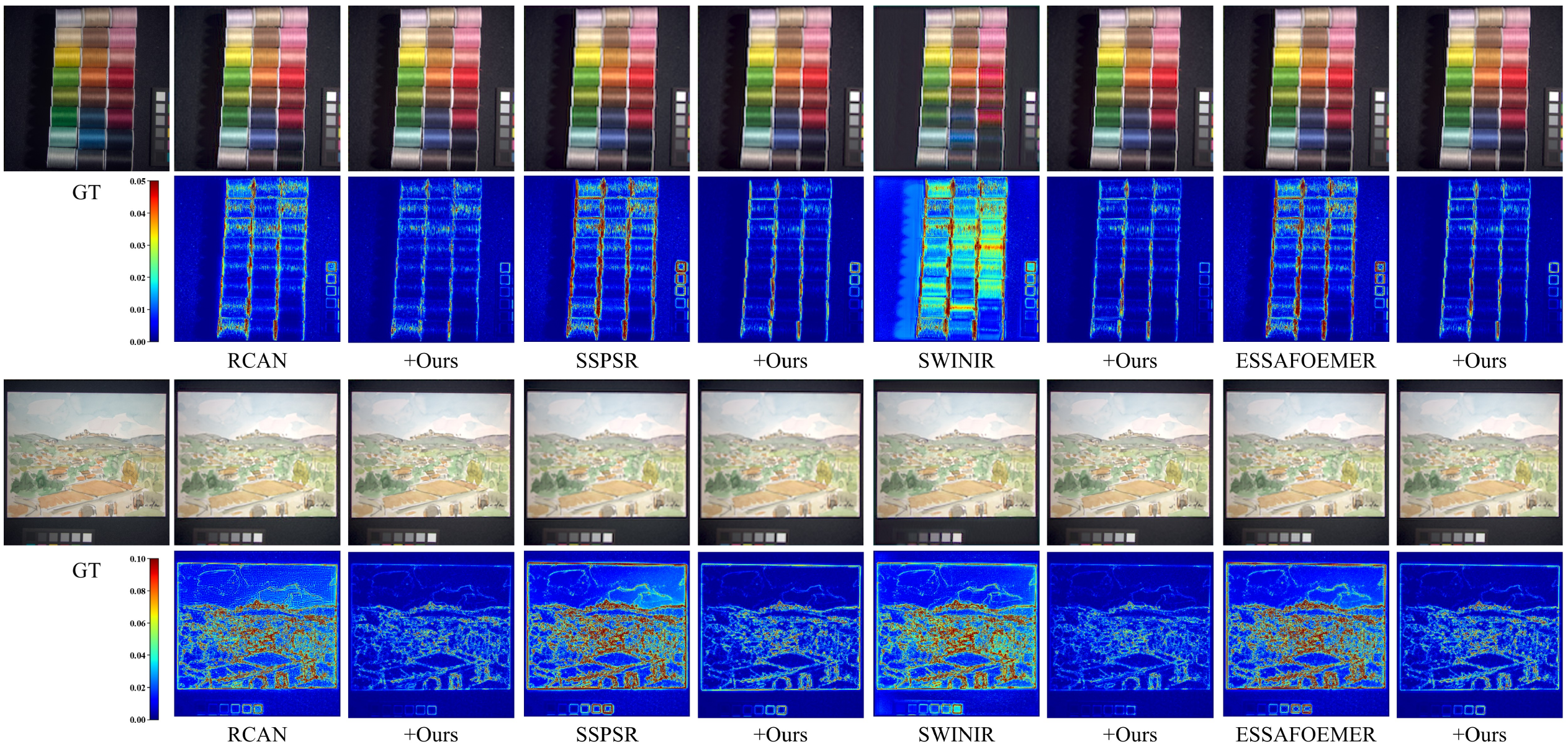}{fig:errmap_cave}{0.31}

\clearpage
\ErrMapGalleryBlock
{Error-map gallery on ICVL at $\times4$ under the cross-domain setting (trained on ARAD-1K, tested on ICVL without fine-tuning). The visualization protocol is identical to Figure~\ref{fig:errmap_arad}.}
{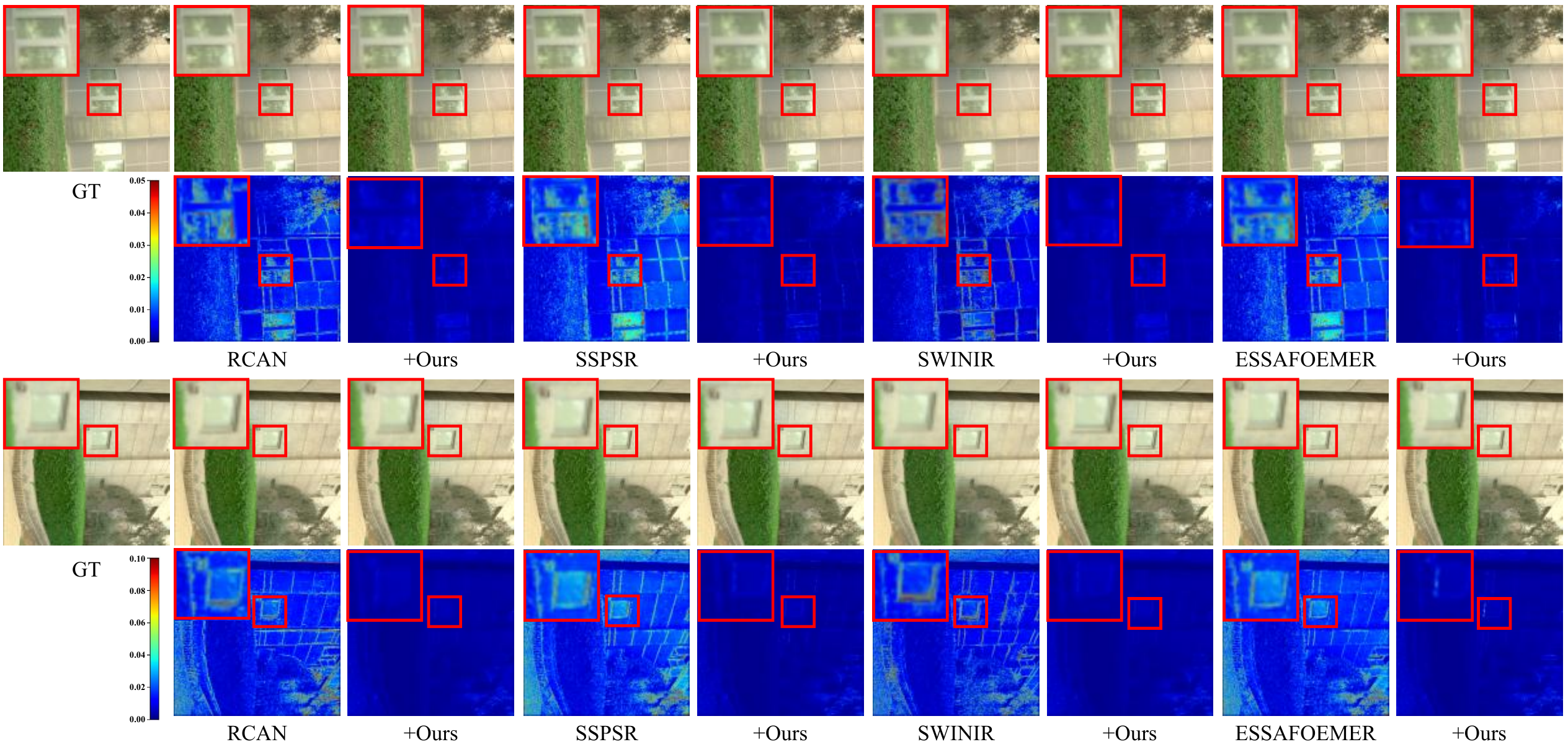}{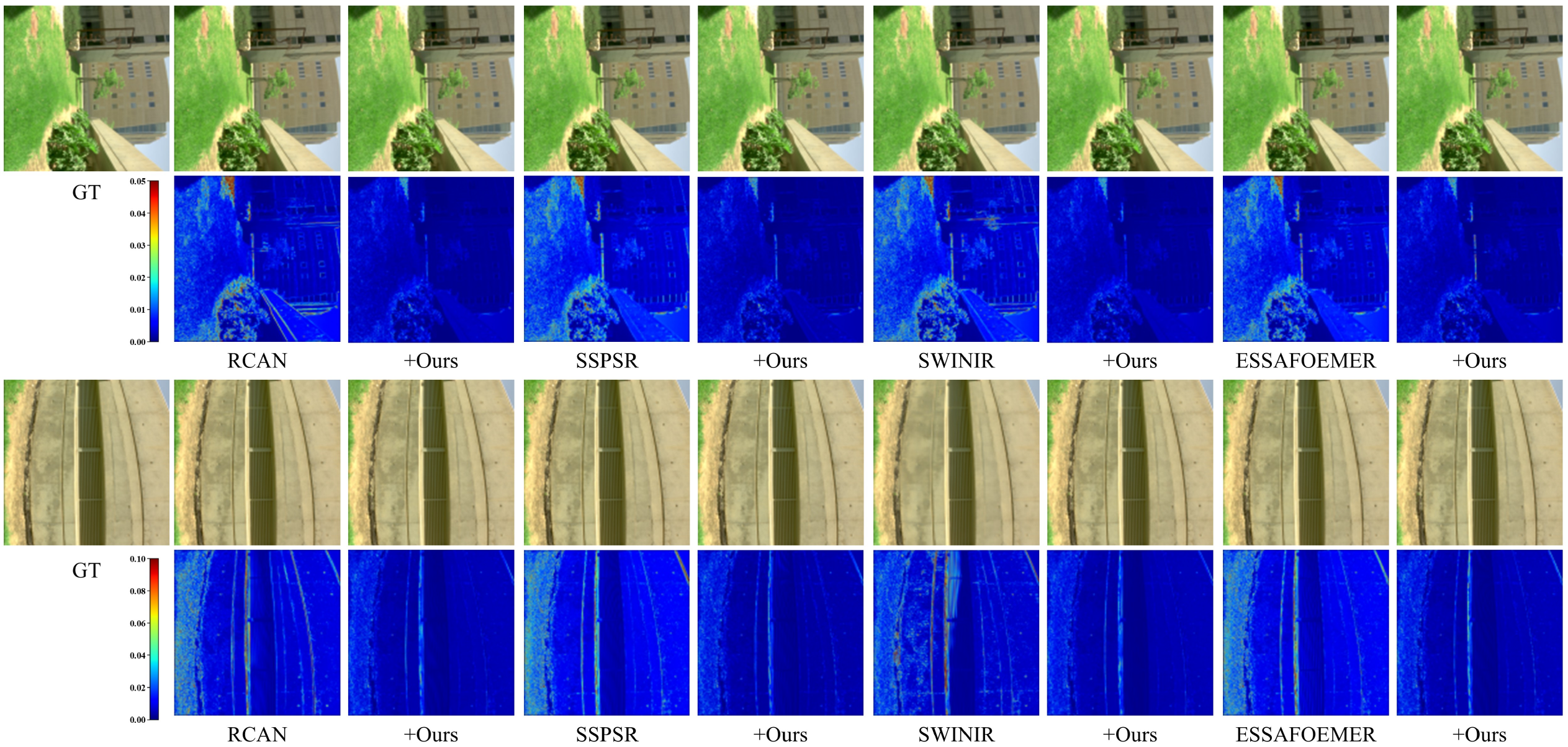}{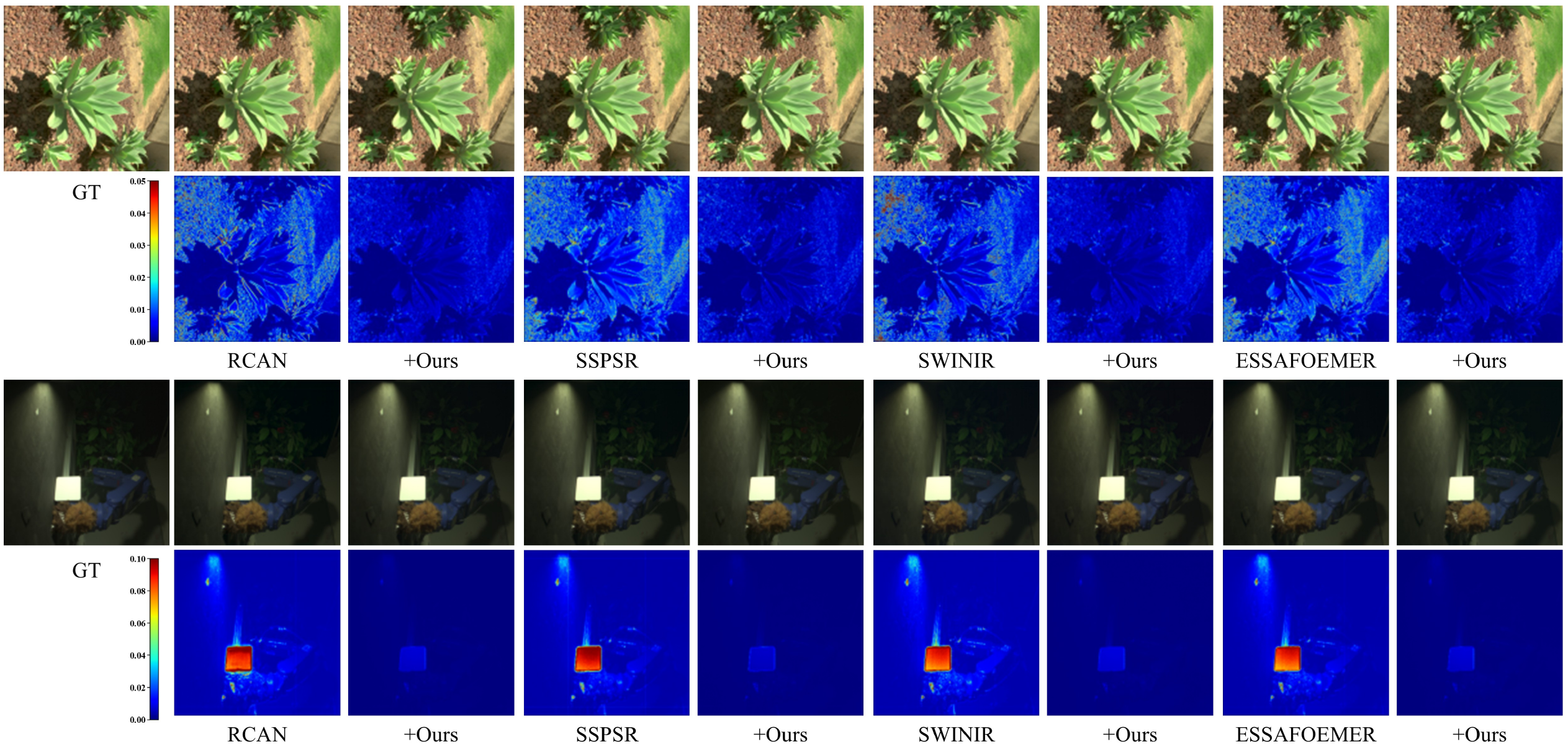}{fig:errmap_icvl}{0.30}

\clearpage
\section{Additional Spectral-Curve Galleries}
\label{sec:supp_spectral}
Figures~\ref{fig:sp_cave_gallery}--\ref{fig:sp_arad_gallery} provide additional per-pixel spectral-curve galleries on CAVE, ICVL, and ARAD-1K at $\times4$, spanning CNN-, Transformer-, and diffusion-based backbones (including DDS2M). For each marked pixel,~\name (green dashed) follows the ground-truth spectrum (blue) more closely \mbox{than the baseline (orange).}

\noindent\begin{minipage}{\textwidth}
\centering
\captionsetup{skip=3pt,hypcap=false}
\includegraphics[width=0.99\linewidth,height=0.255\textheight,keepaspectratio]{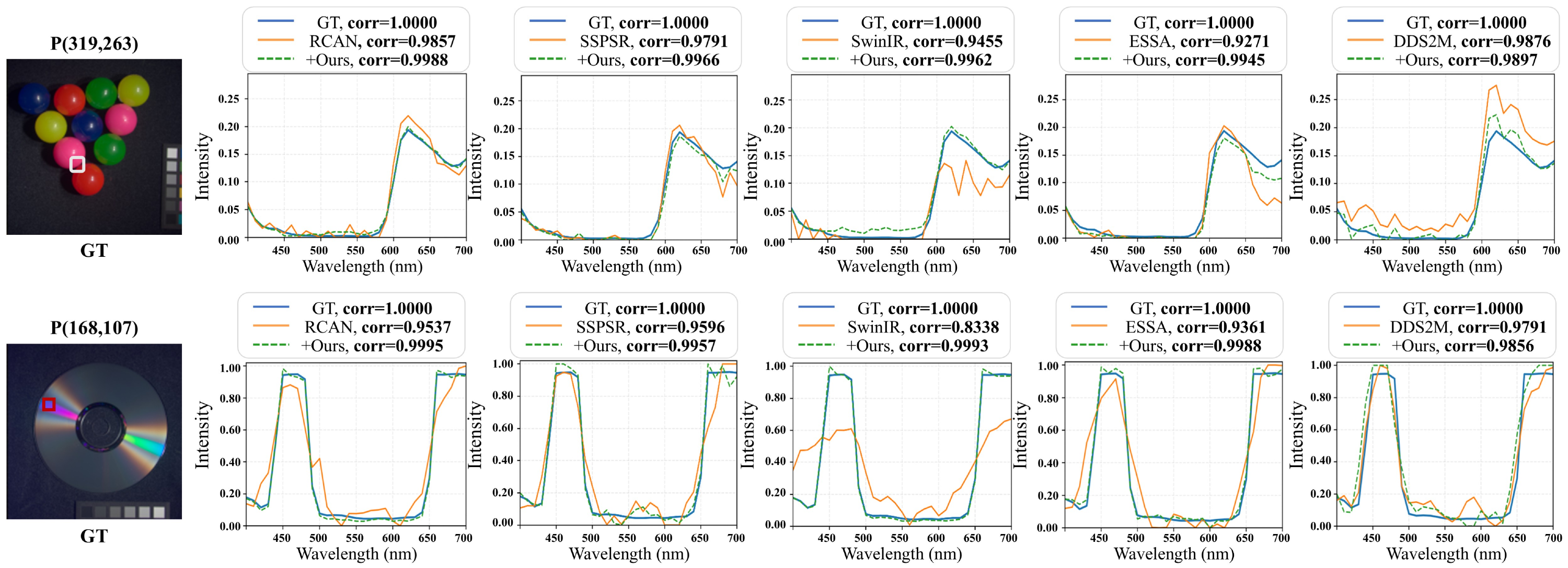}\\[4pt]
\includegraphics[width=0.99\linewidth,height=0.255\textheight,keepaspectratio]{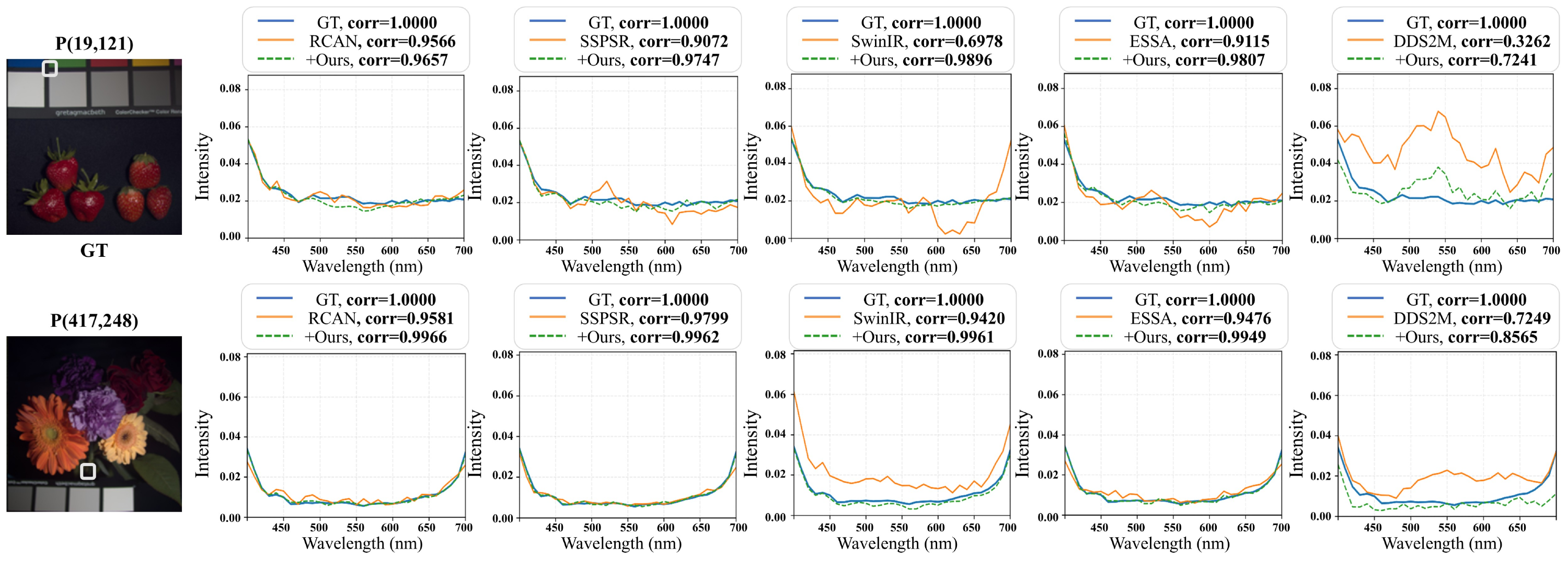}\\[4pt]
\includegraphics[width=0.99\linewidth,height=0.255\textheight,keepaspectratio]{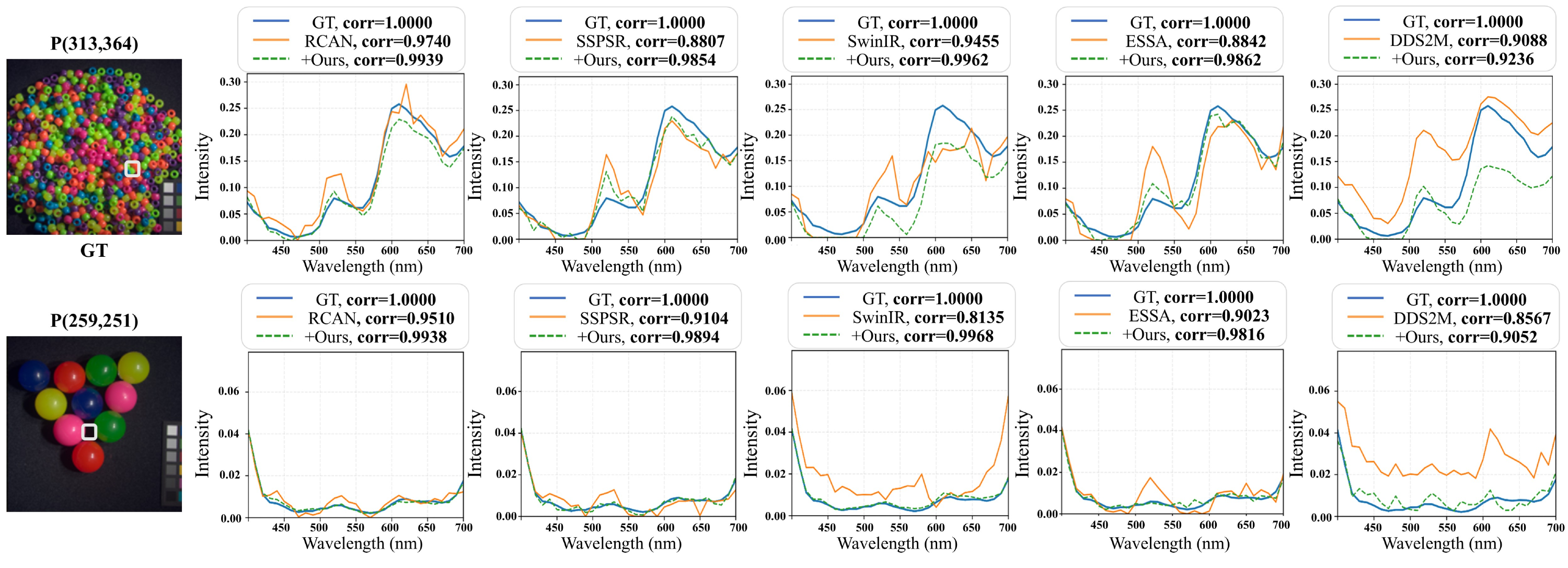}
\captionof{figure}{Additional per-pixel spectral-curve gallery on CAVE at $\times4$.}
\label{fig:sp_cave_gallery}
\end{minipage}

\clearpage
\noindent\begin{minipage}{\textwidth}
\centering
\captionsetup{skip=3pt,hypcap=false}
\includegraphics[width=0.99\linewidth,height=0.30\textheight,keepaspectratio]{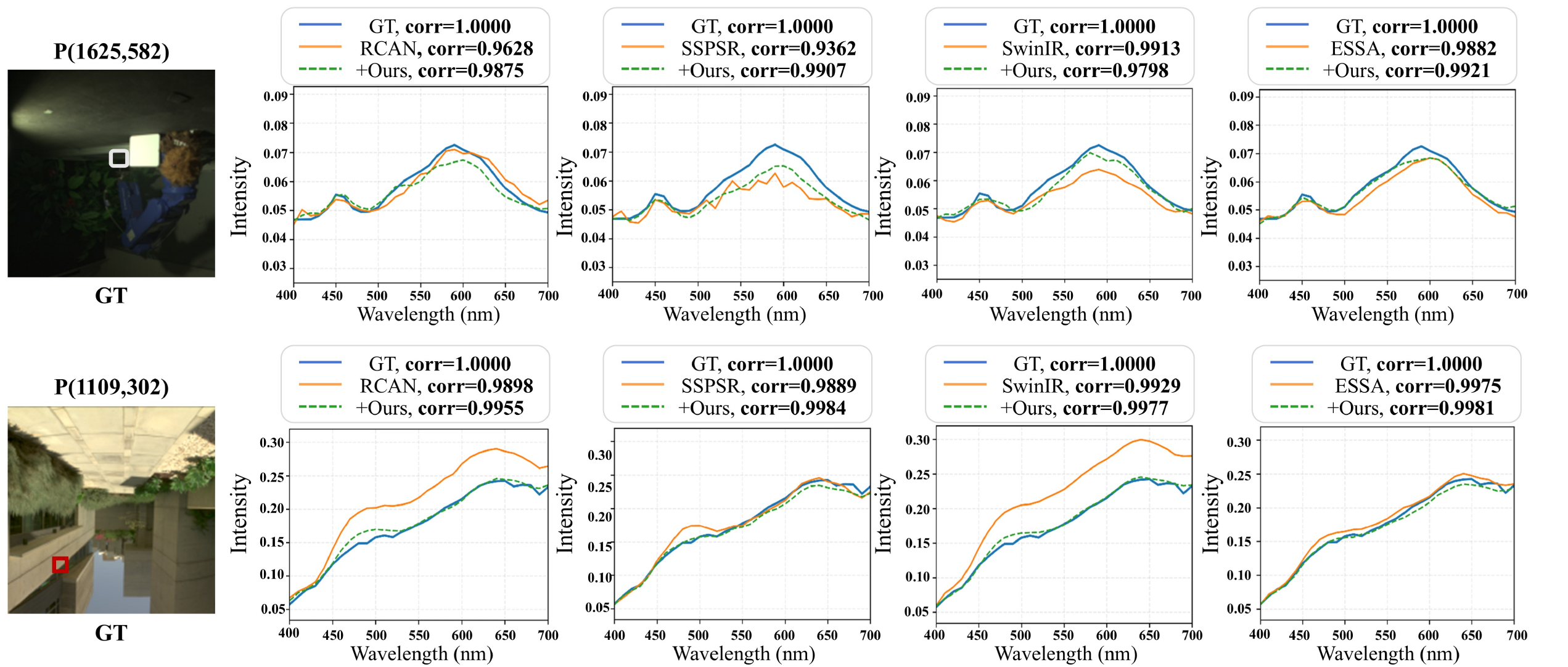}\\[6pt]
\includegraphics[width=0.99\linewidth,height=0.30\textheight,keepaspectratio]{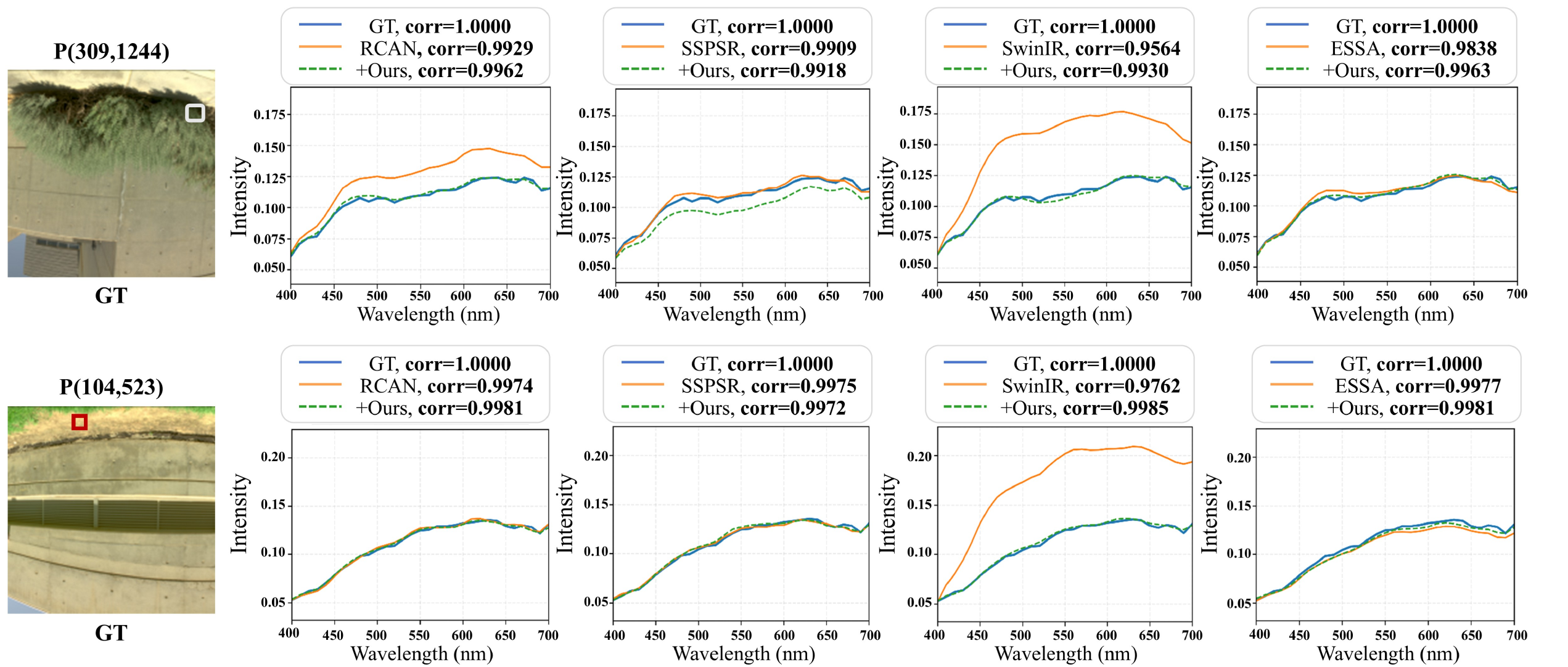}\\[6pt]
\includegraphics[width=0.99\linewidth,height=0.30\textheight,keepaspectratio]{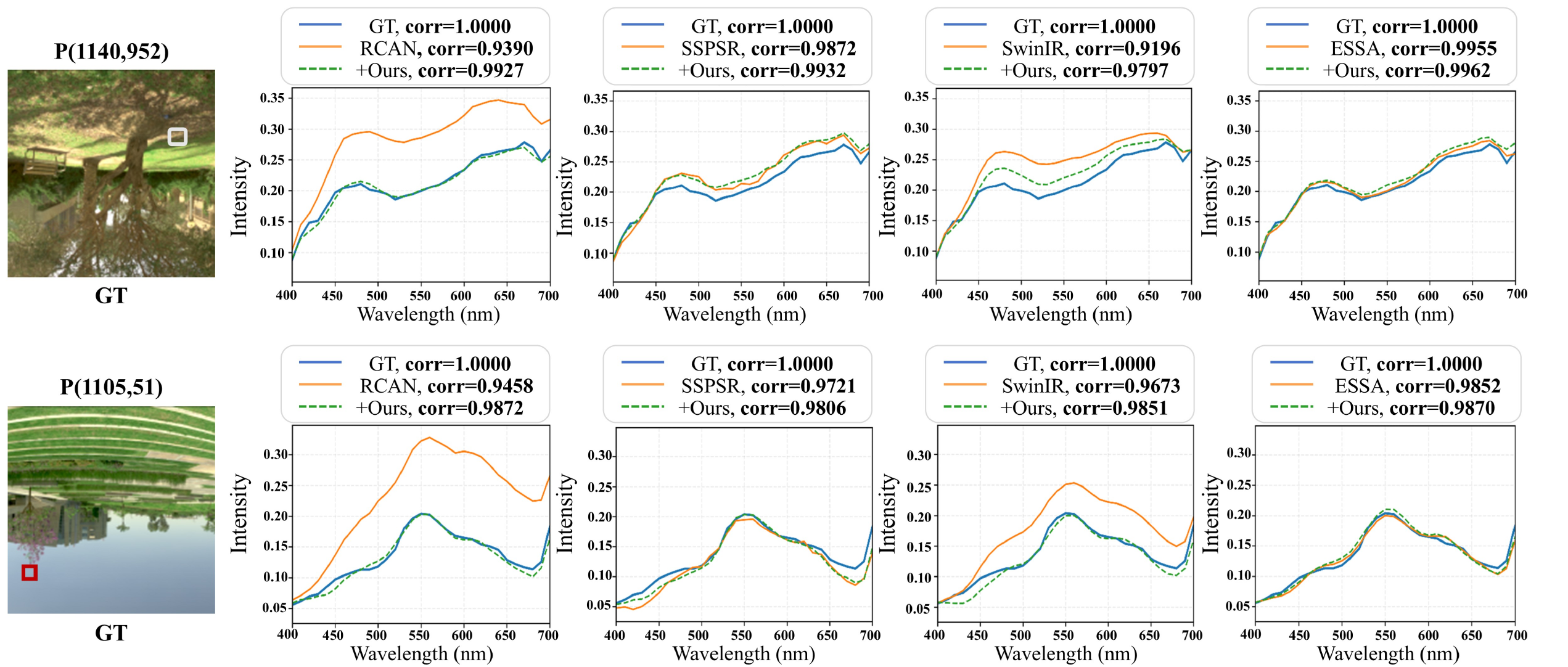}
\captionof{figure}{Additional per-pixel spectral-curve gallery on ICVL at $\times4$ under the cross-domain setting.}
\label{fig:sp_icvl_gallery}
\end{minipage}

\clearpage
\noindent\begin{minipage}{\textwidth}
\centering
\captionsetup{skip=3pt,hypcap=false}
\includegraphics[width=0.99\linewidth,height=0.30\textheight,keepaspectratio]{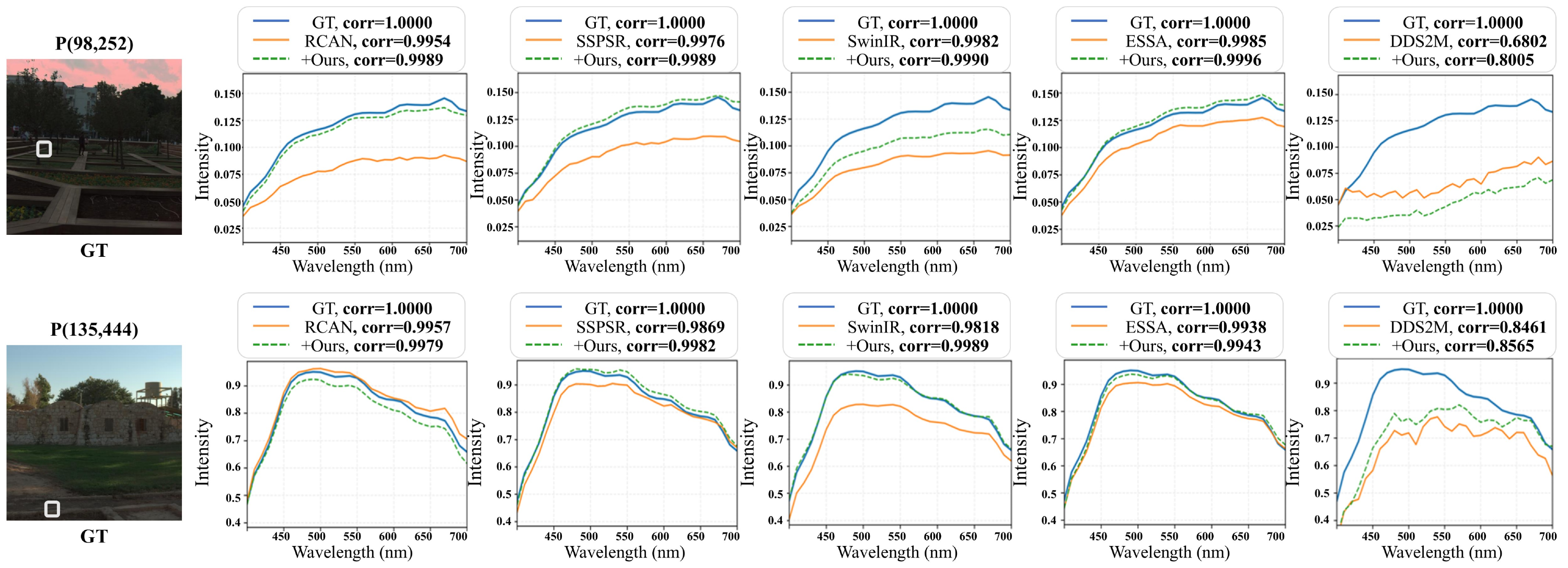}\\[6pt]
\includegraphics[width=0.99\linewidth,height=0.30\textheight,keepaspectratio]{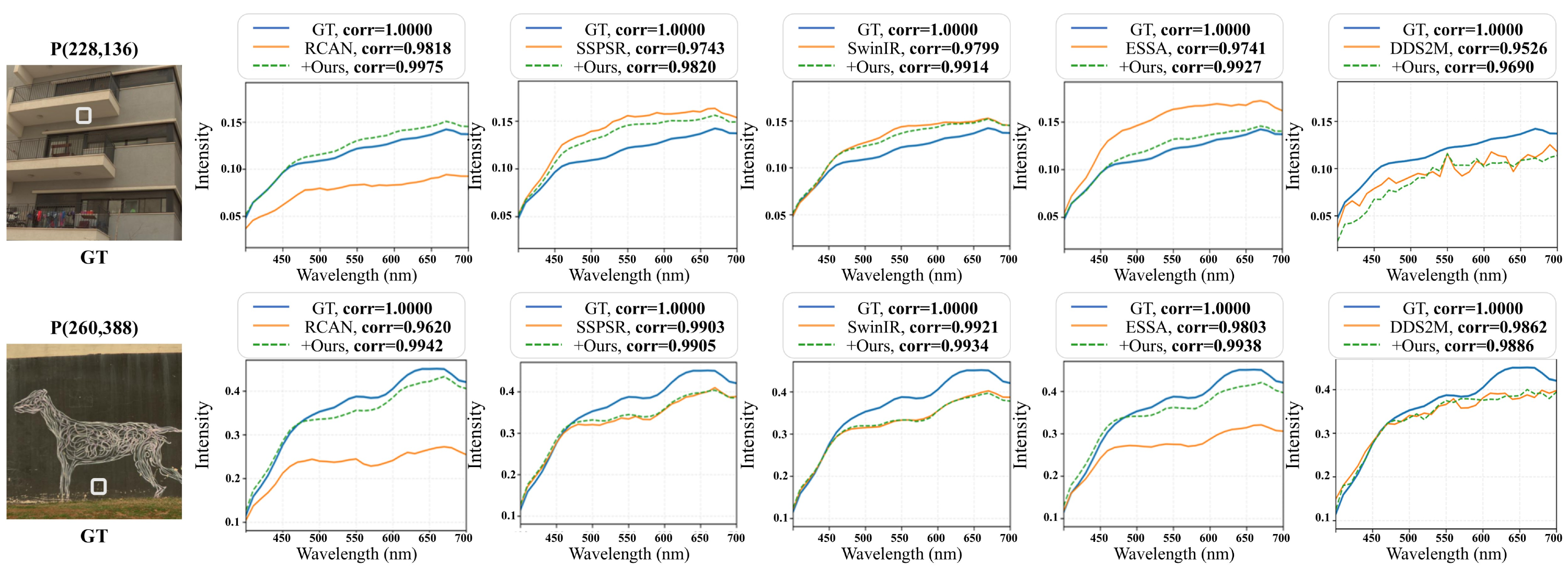}\\[6pt]
\includegraphics[width=0.99\linewidth,height=0.30\textheight,keepaspectratio]{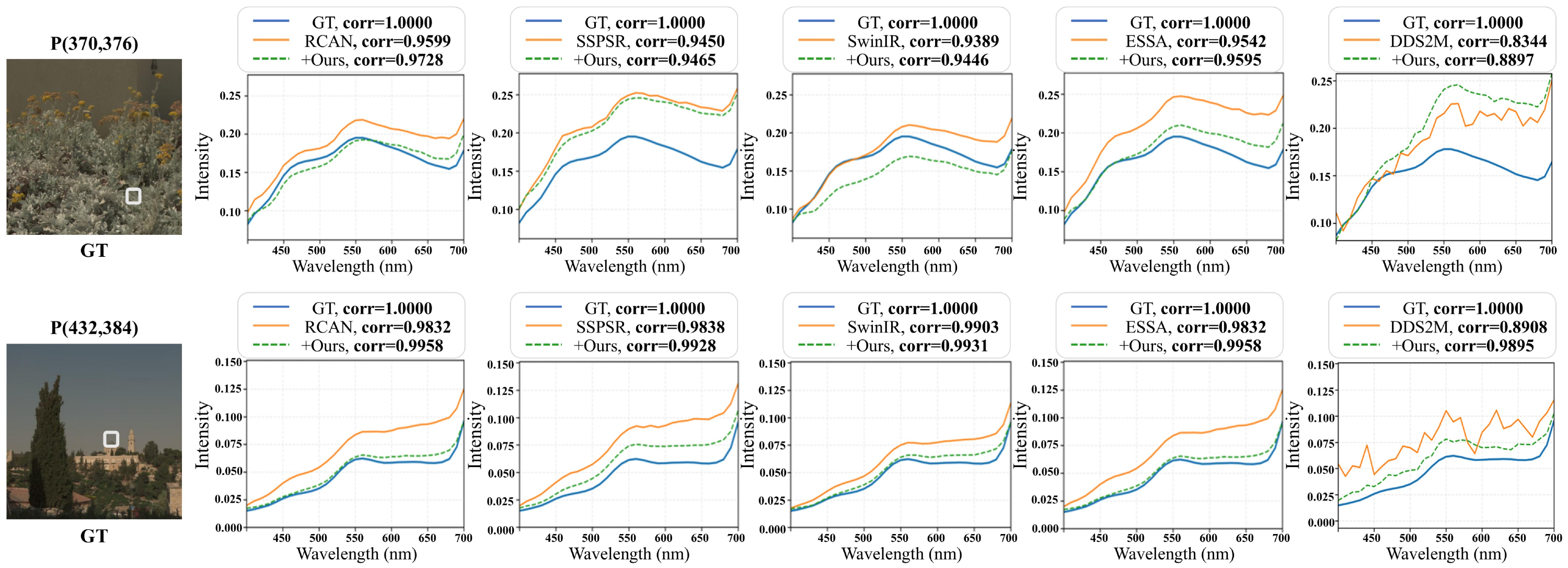}
\captionof{figure}{Additional per-pixel spectral-curve gallery on ARAD-1K at $\times4$.}
\label{fig:sp_arad_gallery}
\end{minipage}

\end{document}